\documentclass[sn-basic,iicol]{sn-jnl}

\usepackage[protrusion=true,expansion=true]{microtype}
\usepackage{hyperref}
\usepackage{array}

\usepackage[normalem]{ulem}

\hypersetup{
  pdftitle={Visuo-Haptic Object Perception for Robots: An Overview},
  pdfauthor={Navarro-Guerrero, Toprak, Josifovski, and Jamone},
  pdfsubject={Robotics (cs.RO); Artificial Intelligence (cs.AI)},
  pdfkeywords={Tactile Sensing, Haptics, Robot Perception, Sensor Fusion, Object Manipulation, Multimodal Machine Learning},%
}

\jyear{2022}%

\begin{document}

\title[Visuo-Haptic Object Perception for Robots]{Visuo-Haptic Object Perception for Robots: An Overview}


\author*[1]{\fnm{Nicol\'as} \sur{Navarro-Guerrero}}\email{nicolas.navarro.guerrero@gmail.com}

\author{\fnm{Sibel} \sur{Toprak}}\email{sibel.toprak@outlook.com}

\author[2]{\fnm{Josip} \sur{Josifovski}}\email{josip.josifovski@tum.de}

\author[3]{\fnm{Lorenzo} \sur{Jamone}}\email{l.jamone@qmul.ac.uk}

\affil*[1]{\orgdiv{Robotics Innovation Center}, \orgname{Deutsches Forschungszentrum f\"ur K\"unstliche Intelligenz (DFKI) GmbH}, \orgaddress{\street{Robert-Hooke-Stra\ss e 1}, \city{Bremen}, \postcode{28359}, \state{Bremen}, \country{Germany}}}

\affil[2]{\orgdiv{School of Computation, Information and Technology}, \orgname{Technische Universit\"at M\"unchen}, \orgaddress{\street{Arcisstra\ss e 21}, \city{Munich}, \postcode{80333}, \state{Bavaria}, \country{Germany}}}

\affil[3]{\orgdiv{Advanced Robotics at Queen Mary (ARQ), School of Engineering and Materials Science}, \orgname{Queen Mary University of London}, \orgaddress{\street{Mile End Road}, \city{London}, \postcode{E1 4NS}, \state{England}, \country{United Kingdom}}}

\abstract{The object perception capabilities of humans are impressive, and this becomes even more evident when trying to develop solutions with a similar proficiency in autonomous robots. While there have been notable advancements in the technologies for artificial vision and touch, the effective integration of these two sensory modalities in robotic applications still needs to be improved, and several open challenges exist. Taking inspiration from how humans combine visual and haptic perception to perceive object properties and drive the execution of manual tasks, this article summarises the current state of the art of visuo-haptic object perception in robots. Firstly, the biological basis of human multimodal object perception is outlined. Then, the latest advances in sensing technologies and data collection strategies for robots are discussed. Next, an overview of the main computational techniques is presented, highlighting the main challenges of multimodal machine learning and presenting a few representative articles in the areas of robotic object recognition, peripersonal space representation and manipulation. Finally, informed by the latest advancements and open challenges, this article outlines promising new research directions.}

\maketitle


\section{Introduction}
\label{sec:intro}
In humans, vision is the most important source of information for object perception. However, haptic feedback is crucial, too. The challenges posed by the absence of vision can be easily experienced by anyone just by trying to perform daily tasks blindfolded or in the dark. Less common is to experience the lack of haptic perception. Frigid fingers, caused by either coldness (e.g., frostnip or frostbite) or specific health conditions (e.g., anaemia), are one example; simply wearing thick gloves is another one, although the impairment is less evident. Early scientific experiments conducted by \cite{Westling1984Factors} have shown how simple manipulation tasks, such as lighting a match, become almost impossible if the tactile feedback is removed by temporarily anaesthetizing the fingertips.

The situation is similar for robots. While vision is a primary source of information, some important object properties cannot be perceived using (only) vision, such as weight, material, or texture. Imagine the case of a robot sorting boxes based on whether they are empty or not without inspecting their content. Such a robot can only do this job if it can perceive the weight of the boxes, both to adjust the grip force (also combining the perceived friction coefficient, i.e., by feeling the texture) and to correctly classify the boxes.
In addition, even for properties that are well detected by vision, such as the position or shape of the object, there are cases in which the sole reliance on this sensory modality is limiting, for example in settings characterized by unpredictable changes in the lighting conditions, or when dealing with translucent, reflecting, and occluded objects.
Relying on multiple sensory modalities can help resolve these perceptual ambiguities. 

The idea of integrating vision and touch was first proposed by \cite{Allen1984Surface} to generate descriptions of object surfaces. \cite{Allen1988Integrating} extended this idea to encompass the whole object recognition task. Since then, much work has been done on recognizing and manipulating objects based on one modality, i.e., based on either vision or haptics alone (Please refer to, e.g., \cite{Zhao2019Object, Fanello2017Visual, Guo2016Deep, Du2021VisionBased} for an extensive overview of visual object perception and \cite{Seminara2019Active, Luo2017Robotic, Kappassov2015Tactile} for an extensive overview of haptic object perception). Despite the significant progress achieved in the field based on either visual or haptic information, the combination thereof has attracted less attention in comparison, e.g., \cite{Liu2017Recent, Yang2015Object}.

Usually, in machine learning applications, visual and haptic perception are treated as two separate processes that converge at some point to a final classification result, e.g., \cite{Liu2020Audiovisual, Cui2020SelfAttention}. However, in the brain, interactions between vision and touch take place in the cerebral cortex \citep{Lacey2016Crossmodal}. These interactions can be crossmodal, meaning that the haptic stimuli activate regions traditionally believed to be visual or multimodal, in which case the visual and the haptic stimuli converge.

This article presents a holistic overview of multimodal object perception for robots from both a bio-inspired and a technical point of view.
Firstly, the biological basis of visuo-haptic object perception is introduced. Secondly, a summary of tactile sensors and multimodal datasets are provided. Thirdly, the computational challenges of multimodal signal processing are presented. Fourth, the main application areas are introduced and reviewed, including multimodal object recognition, peripersonal space representation, and object manipulation. Finally, challenges and future directions for research on artificial visuo-haptic object perception are discussed.

\section{Neural Basis of Visuo-Haptic Object Perception}
\label{sec:bio}

The fact that there is no learning algorithm yet that reaches the level of proficiency of the human brain when it comes to recognizing objects illustrates how complex this cognitive task actually is \citep{Smith2018Developing, Krueger2013Deep, James2007Neural}. The human brain is capable of performing it both quickly and accurately, even when the visual information available is incomplete or ambiguous. One reason might be that the brain can complement that `picture' with information from other sensory modalities at will; usually, it does this with haptics. However, it is also because the learning machinery in the human brain seems to be suited to learn from drastically different frequency distributions than those used in machine learning, as described by \citet{Smith2018Developing}. 
In particular, infants seem to use curriculum learning constrained by their developing sensorimotor abilities and actions. However, what is in strong contrast with machine learning algorithms is that the learning machinery, at least in infants, is particularly effective in learning from extremely skewed frequency distributions, i.e., a very small number of instances are highly frequent while most other instances are encountered very rarely. For instance, in very young infants, more than 80\% of faces they are exposed to are from 2-3 individuals \citep{Smith2018Developing}.

We argue that taking inspiration from the complementary nature of the sensory modalities as well as processes in the brain that are involved in fusing the information they provide during object perception, might help build better robotic systems. While this topic is an active area of research and considerable new insights have been gained, there are still many aspects about the inner workings of the human brain during object perception that are not fully understood. 

In this section, we present a short review of what is known on visuo-haptic object perception and recognition in the brain (or more specifically in the cerebral cortex), focusing on the main organizational and functional principles that can serve as a basis for computational modelling given the complexity of this topic and the abundance of research available.

\subsection{Visual Object Perception}
\label{sec:bio:vision}

For every basic sense, a primary sensory area can be identified in the cerebral cortex, the earliest cortical area in the brain's outer layer to process the sensory stimuli coming from the respective receptors. For vision, that area, the primary visual cortex (V1) \citep{Krueger2013Deep, Grill-Spector2004Human, Malach1995ObjectRelated} is located on the backside of the brain, in what is referred to as the occipital lobe.

The neurons here are organized in a way that allows for neighbouring regions in the retina, and hence in the visual input, to be projected onto neighbouring areas in V1. Retinotopic maps emerge from this orderly arrangement in V1 and subsequent lower visual areas, where the output of the processing at the level of very primitive visual features is forwarded to.

The hierarchical organization of the visual cortical areas and the receptive field size of the neurons gradually increasing with each new area along this hierarchy turns the visual information into more complex and abstract representations \citep{Ungerleider1994what, Krueger2013Deep, Grill-Spector2004Human}. This hierarchical organization is what convolutional neural networks (CNNs) take their inspiration from computationally \citep{Fukushima1980Neocognitron, LeCun2015Deep}.

\begin{figure}[htb]
  \centering
  \includegraphics[width=\columnwidth]{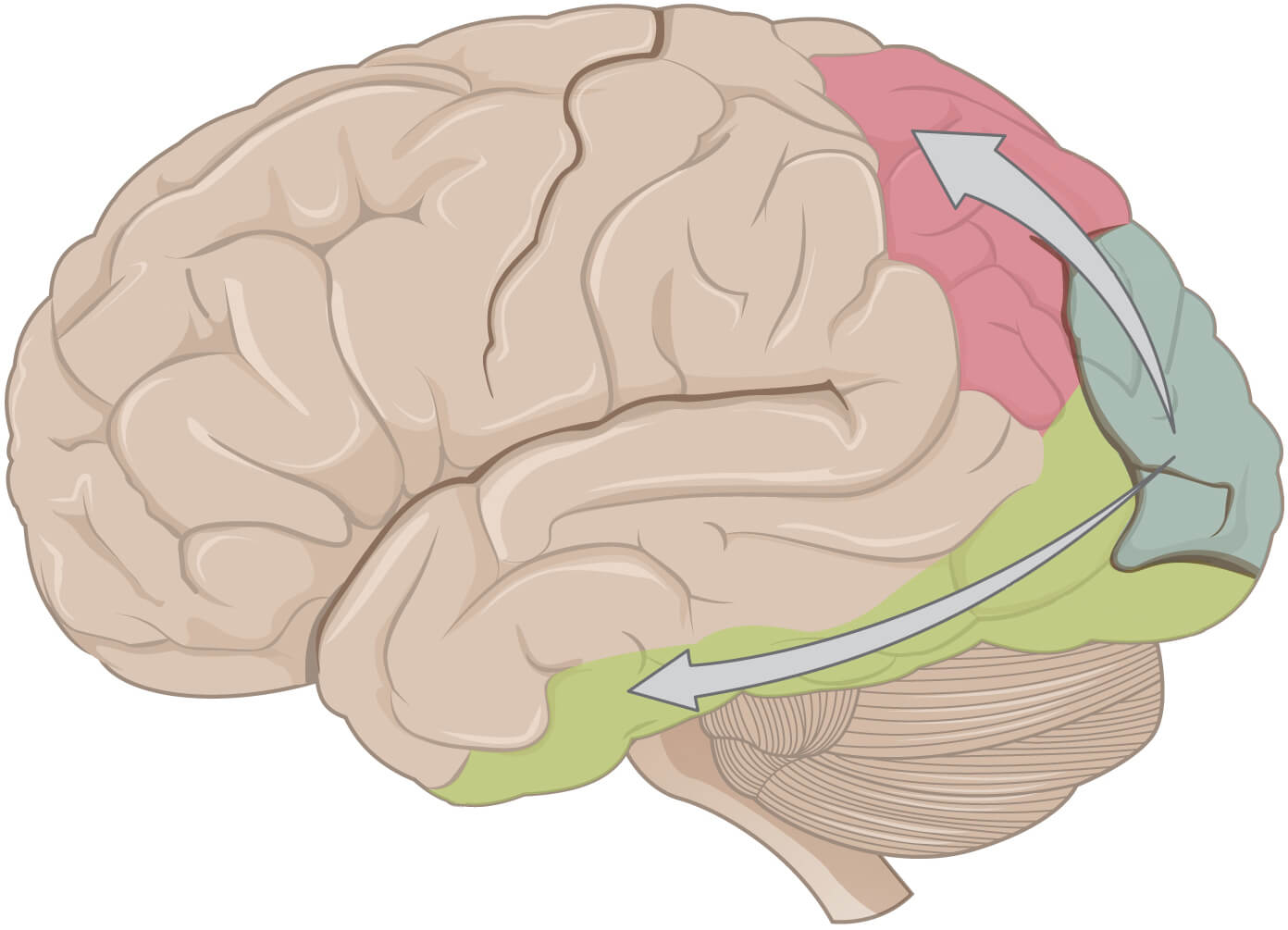}
  \caption{The dorsal and ventral streams originate from the primary visual cortex (V1). The arrow from the right to the top left represents the dorsal stream, and the arrow from the right to the bottom left represents the ventral stream. Adapted from \cite{Young2013Anatomy} CC BY 4.0.}
  \label{fig:ventral-dorsal-model}
\end{figure}

Hierarchical organization aside, the processing of the visual stimuli following V1 has been found to diverge into two main pathways or streams \citep{Ungerleider1994what, Mishkin1983Object}, see Figure \ref{fig:ventral-dorsal-model}. One stream runs ventrally, extending into the temporal lobe of the cortex, and is responsible for the visual identification of objects, while the other runs dorsally, reaching into the parietal lobe, and enables the visual location of and spatial relations among objects \citep{Mishkin1983Object}. The ventral and dorsal streams are, therefore, also called the ``what" and ``where" pathway, respectively. A modification to this model was later introduced to distinguish between ``vision for perception" and ``vision for action" and to emphasize that the dorsal stream also coordinates visually guided actions directed at objects \citep{Goodale1992Separate}. Hence, these streams are alternatively referred to as ``perception" and ``action" pathways. The overall model became known as the \textit{two visual systems (TVS) model} \citep{Rossetti2017Rise, Milner2017How, DeHaan2018Where, Goodale2018Two}.

The idea that the neural substrates underlying each visual processing stream are distinct was initially proposed by \citet{Goodale1991Neurological, Goodale1992Separate} and has been widely accepted since. However, it has become the subject of controversy as of late for being oversimplified \citep{DeHaan2011Usefulness, Sheth2016Two, Rossetti2017Rise, DeHaan2018Where}. There is evidence for cross-talk between the two streams: ventral to dorsal when information about the object and its qualities is required to plan and fine-tune a grasping action \citep{Perry2014Feature, VanPolanen2015Interactions, Milner2017How}, and dorsal to ventral, when updated grasp-related information helps refine the 3D perception and possibly the internal representation of objects \citep{VanPolanen2015Interactions, Freud2016What, Milner2017How}. Nevertheless, the TVS model has inspired a considerable amount of research in this area and hence remains influential \citep{DeHaan2018Where, Goodale2018Two}.

Zooming in on the perception pathway, the division into functional streams seems to be a recurring pattern in the cortex as evidence suggests that there is a further specialization into sub-streams here, one dedicated to object form and another to surface properties \citep{Cant2009fMRAdaptation, Cant2007Attention}. The posterior-lateral regions of the occipito-temporal part of the cerebral cortex, including the lateral occipital area (LO), were shown to contribute to the perception of object form. Meanwhile, the more medial parts of the ventral stream handle the perception of object surface properties like texture or colour. In particular, areas along the collateral sulcus (CoS) have been found to respond to texture specifically. In contrast, an analogous area for colour could not be identified: it is believed that the processing of information related to surface colour occurs relatively early along the ventral stream compared to surface texture.
In general, it appears that areas showing form selectivity overlap with those involved in object recognition and identification. Similarly, there seems to be an overlap between areas selective to object surface properties with the fusiform gyrus (FG), an area in the temporal lobe taking care of perception of more complex stimuli categories like faces and places \citep{Cant2007Attention}.

Further studies have confirmed and added to these findings \citep{Cavina-Pratesi2010Separatea, Cavina-Pratesi2010Separate}. Accordingly, there is not one single cortical area but multiple interacting foci in the medial ventral stream region that infer the material properties of perceived objects from extracted individual surface properties. A texture-selective area appears to be located posterior to a colour-selective one. Also, areas showing responsiveness to multiple object properties were detected next to areas of dedicated single-feature processing \citep{Cavina-Pratesi2010Separatea, Cavina-Pratesi2010Separate}.

Overall, visual information can be located at three different levels of abstractions in the cerebral cortex along the ventral visual stream: between retinotopy and stimulus categories (objects, faces, places, etc.), there is an intermediate level of representation based on geometric and material properties \citep{Cavina-Pratesi2010Separatea}.
This hierarchical functional organization is advantageous \citep{Krueger2013Deep}: using separate but highly interconnected channels for processing different types of visual information (colour, shape, etc.) allows for representations that are both robust against missing cues and efficient, as the combinatorial explosion and the resulting lack of generalization to new objects that an integrated representation would cause, is prevented.

\subsection{Prehension of Objects}
\label{sec:bio:haptics:prehension}


Object perception benefits greatly from performing exploratory procedures (EPs) on an object of interest, to observe different sides of an object or perceive non-visual features for instance. For that, we first reach towards that object, i.e., move our hand close to its location, and then grasp it, which involves pre-shaping our hand to the object's physical properties and selecting the optimal grip type. The capacity to reach and grasp objects is also more generally referred to as prehension \citep{Turella2014Neural}.

Initially, it was thought that the detailed organization of the dorsal stream reflects these two components of prehension, again in the form of independent pathways, as in the case of the ventral stream (see Section \ref{sec:bio:vision}). According to this classical model, one pathway comprises the more laterally located areas of the dorsal stream and controls grasping, whereas the medial areas form the other pathway, which is recruited during reaching. Hence, these two pathways are also called the dorsolateral and dorsomedial pathways, respectively \citep{Fattori2010Dorsomedial, Turella2014Neural, Rizzolatti2003Two}.

Later on, it was shown that this initial model has limitations: \cite{Fattori2010Dorsomedial}, for instance, offers evidence that the dorsomedial pathway is not only for reaching and that it may play a central part in all phases of reach-to-grasp action. In their review on the coding of prehension in the brain, \cite{Turella2014Neural} conclude that the coding of grasping, maybe even the integration with reaching, seems to happen in both pathways and that the temporal difference in the onset of processing suggests that the processing in the dorsomedial pathway is driven by the dorsolateral one. The authors argue that this aspect could yield a more fitting functional characterization of the pathways instead of grasping and reaching: There is strong evidence that the dorsolateral pathway is in charge of creating an action plan and the dorsomedial one follows with online adjustment.

More recent findings support that the role of the dorsomedial pathway goes beyond just online control and adjustment during prehension: It has been suggested that the early dorsomedial areas are involved in the biomechanical selection of viable grasp postures during reach-to-grasp behaviours \citep{Galletti2018Dorsal} and even before, that is in preparation of the action execution \citep{Santandrea2018Preparatory}.

\subsection{Importance of Haptics for Object Perception}
\label{sec:bio:haptics}
Although we primarily rely on our vision for object perception and recognition, we may occasionally use our other senses in the face of very ambiguous, and hence difficult, cases. The sensory modality that we then typically resort to is haptics, which is complementary to vision in many regards. With our vision, we are capable of perceiving multiple object properties at one glance, whereas haptic perception can involve a sequence of steps to accomplish the same \citep{Lederman1987Hand}. Our eyes may sometimes provide access to only a limited perceptual space, be it due to visual impairments or the conditions in our environment. In such cases, our skin, as our largest sensory organ, combined with active touch and exploration, can help us enlarge that space and perceive what we otherwise would not be able to. That is because the sets of visually and haptically perceivable object properties are largely complementary.

\begin{figure}[tb]
  \centering
  \includegraphics[height=0.12\textheight]{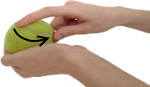}\hfill
  \includegraphics[height=0.12\textheight]{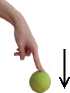}
  \includegraphics[height=0.12\textheight]{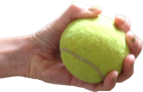}\hfill
  \includegraphics[height=0.12\textheight]{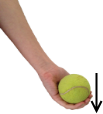}
  \includegraphics[height=0.117\textheight]{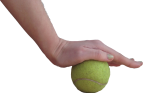}\hfill
  \includegraphics[height=0.117\textheight]{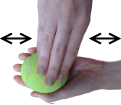}
  \caption{Illustration of six exploratory procedures, as described by \citet{Lederman2009Haptic}. From left to right and top to bottom: Contour Following, Pressure, Enclosure, Unsupported Holding, Static Contact, and Lateral Motion. Adapted from \cite{Nelinger2015Tactile} CC BY 3.0.}
  \label{fig:exploratory-procedures}
\end{figure}

\cite{Lederman1987Hand} have identified patterns for how objects are typically explored manually. These patterns are referred to as exploratory procedures (EPs) \citep{Lederman1987Hand, Lederman2009Haptic}. These EPs can be roughly distinguished into three categories, namely those related to the substance of an object (texture, hardness, temperature, and weight), those related to the structural properties of an object (global shape and exact shape, volume, and weight) and those for discovering the function of an object (finding the movable parts, deducing the potential function based on its form). Examples of the exploratory procedures for the first two categories are shown in Figure \ref{fig:exploratory-procedures}.

There are eight EPs in total \citep{Lederman1987Hand}: an object's texture can be explored using the \textit{lateral motion EP}, where the fingers or other parts of the skin are moved along its surface. With the \textit{pressure EP}, which can manifest itself in either a poking or tapping movement, the hardness of an object can be tested. The \textit{static contact EP} is for feeling the object's temperature by briefly and passively touching its surface. Using the \textit{unsupported holding EP}, an object's weight can be inferred from the effort needed to balance the object at a certain height. An object's global shape and volume can be sensed with the help of the \textit{enclose EP}, which involves placing the hands around the object to cover as much of its surface as possible, repeatedly if needed, and positioning the hands differently each time. During the \textit{contour following EP}, the object's contours are traced, which allows for the local shape or volume of an object to be perceived in more detail. The \textit{part motion test EP} is used to detect to which extent object parts move when force is applied to them, while the \textit{function test EP} examines what functions an object can potentially fulfil by randomly interacting with it.

\subsection{Haptic Object Perception}
\label{sec:bio:haptics:perception}
\begin{figure}[htbp]
  \centering
  \includegraphics[width=\columnwidth]{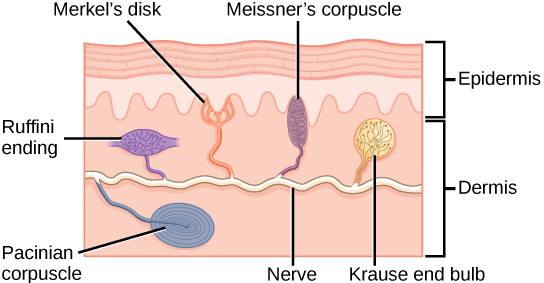}
  \caption{Primary mechanoreceptors in the human skin. Merkel's cells respond to light touch, Meissner's corpuscles respond to touch and low-frequency vibrations. Rufinni endings respond to deformations and warmth. Pacinian corpuscles respond to transient pressure and high-frequency vibrations. Krause end bulbs respond to cold. Image from \cite{Clark2020Biology} CC BY 4.0.}
  \label{fig:skin}
\end{figure}

We usually (and intuitively) think of haptic perception as anything we can perceive using our touch sense, i.e., our skin. The skin is innervated with receptors that can be divided into three groups based on their function \cite[Chap.\ 9]{Purves2012Neuroscience}: mechanoreceptors react to mechanical pressure or vibration and thermoceptors to changes in temperature, whereas nociceptors create the sensation of pain in the case of powerful stimuli that could be damaging, see Figure \ref{fig:skin}.

However, proprioception, the sense of self-movement and body position perceived from stimuli originating from receptors embedded in the muscles, joints, and tendons \citep{Lederman2009Haptic, Dahiya2013Tactile}, often also called kinesthesia, plays an essential role in the haptic perception of objects. An object property that shows the relevance of the kinesthetic sense is shape \citep{Lederman2009Haptic}: what helps us determine an object's shape is the alignment of the bones and the stretching of our muscles when we enclose it with our hands. Similarly, when we are prompted to describe the shape of an object, we tend to demonstrate it with hand poses.

\begin{figure}[htbp]
  \centering
  \includegraphics[width=\columnwidth]{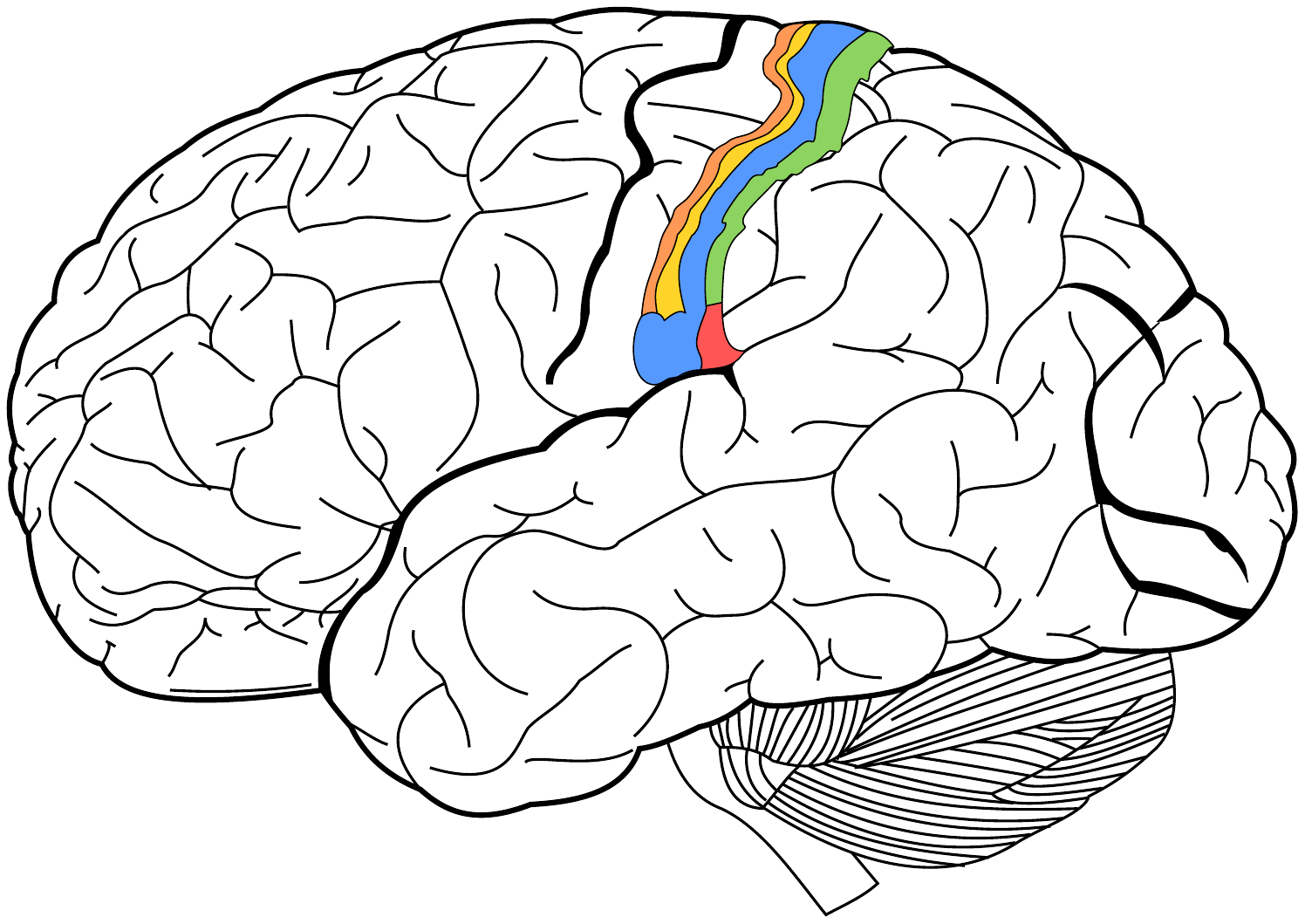}
  \caption{Somatosensory Cortex. The primary somatosensory cortex (S1) consists of the Area 1 (Blue), Area 2 (Green), Area 3a (Orange), and Area 3b (Yellow). The secondary somatosensory cortex (S2) is depicted in red. Image derivative from \href{Selket}{https://en.wikipedia.org/wiki/File:Ventral-dorsal_streams.svg} under CC BY-SA 3.0 and based on \citet[][p., 202]{Purves2012Neuroscience}.}
  \label{fig:somatosensory-cortex}
\end{figure}

\begin{figure}[htbp]
  \centering
  \includegraphics[width=\columnwidth]{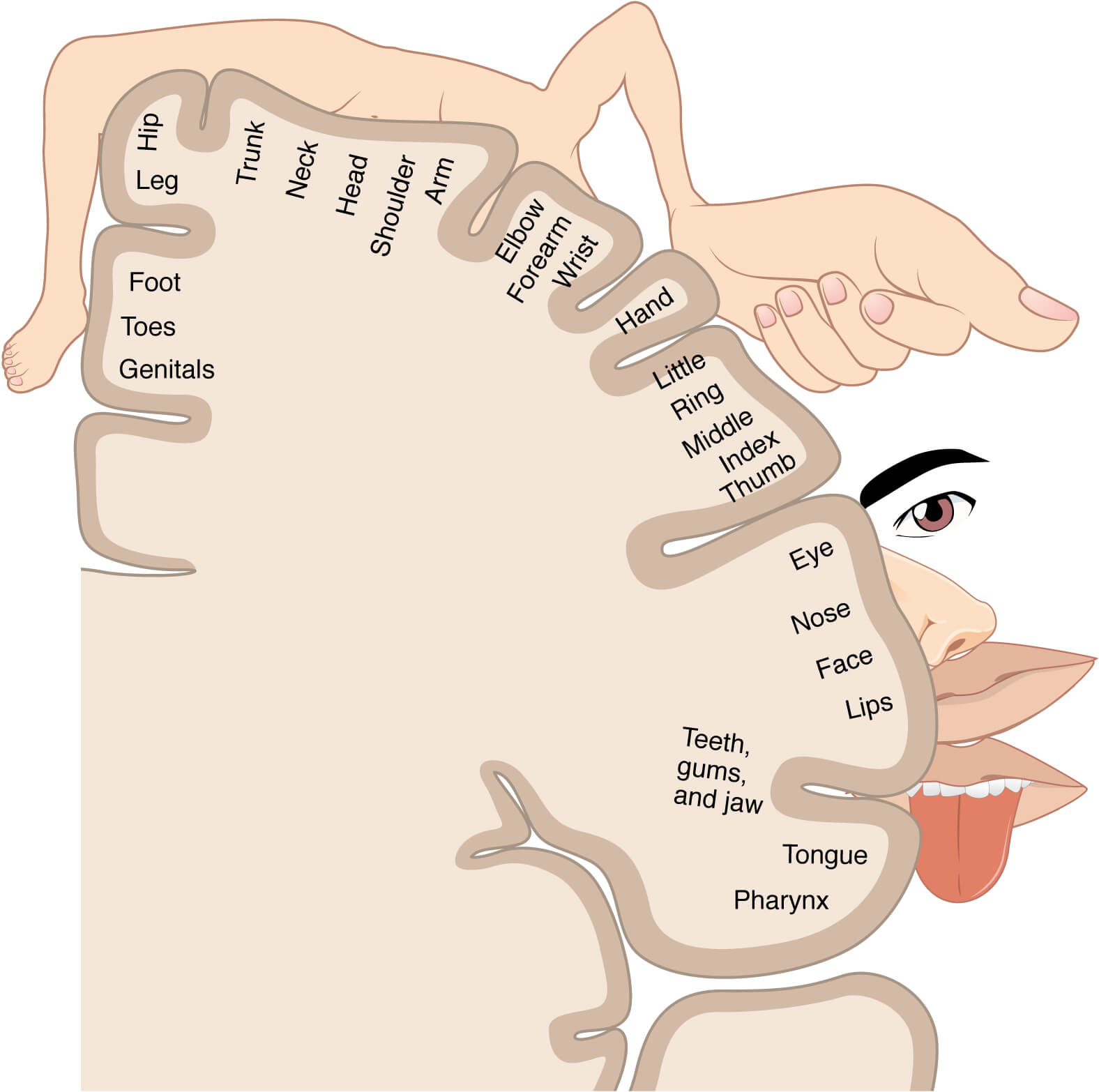}
  \caption{The cortical sensory Homunculus. A representation of the human body based on the proportions of the cortical regions dedicated to processing sensory functions. Image from \cite{Young2013Anatomy} CC BY 4.0.}
  \label{fig:homunculus}
\end{figure}

The primary sensory area for haptic perception is the primary somatosensory cortex (S1) \citep[Chap.\ 9]{Purves2012Neuroscience} \citep{James2007Neural}. It is located in the parietal lobe in the so-called postcentral gyrus and is, from anterior to posterior, comprised of the Brodmann areas 3, further subdivided into 3a and 3b, 1 and 2, see Figure \ref{fig:somatosensory-cortex}. S1 is organized somatotopically across all Brodmann areas. Like retinotopy, somatotopy is a form of topographical organization, resulting in a map of the complete body in each Brodmann area, though not in actual proportion: the area dedicated to each body part in S1 directly reflects the density of receptors in it. The feet, legs, trunk, forelimbs, and face are represented from medial to lateral in these somatotopic maps, see Figure \ref{fig:homunculus}.

Like vision, the processing of the somatic sensations occurs hierarchically: each area receives the information from the periphery, but areas 1 and 2 also receive input from 3a and 3b. Most of the initial processing of the somatosensory input happens in area 3, where area 3a is concerned explicitly with the proprioceptive and 3b with the cutaneous stimuli. Because area 3b is densely connected to areas 1 and 2, the extracted cutaneous information is forwarded to these areas for higher-level processing. Here, area 1 seems to be in charge of texture discrimination, and area 2, involving proprioceptive stimuli, of size and shape discrimination.

The functional divergence into separate pathways might not be only specific to the visual system. The somatosensory system may be organized similarly with two or potentially even more pathways \citep{Sathian2011Dual, James2010Dorsal}, though different views exist on this matter, see \cite{James2010Dorsal} for a review. Object-related haptic activation has been detected outside the somatosensory cortex in multiple areas along the ventral visual pathway. The lateral occipital complex (LOC) was found to respond selectively to object features in both vision and haptics \citep{Malach1995ObjectRelated}. In particular, a subregion of the LOC called lateral occipital tactile-visual region (LOtv) appears to be a bimodal convergence area concerned with the recovery of the geometric shape of objects \citep{Amedi2001VisuoHaptic, Amedi2002Convergence, Tal2009Multisensory}. While not bimodal in nature, haptic activation was also detected in the medial occipitotemporal cortex in response to surface texture \citep{Podrebarac2014Are, Whitaker2008Vision}. This area is close to the one along the CoS concerned with visual texture perception but still spatially distinguishable. The representation of texture information in the visual and haptic modalities differs from that of shape information. However, the processing might not be entirely independent: the proximity of both areas might, in fact, enable cross-modal interaction.

The representation of object weight is located in the medial ventral visual pathway as well \citep{Gallivan2014Representation, Kentridge2014Object}, which might also explain our ability to associate a certain weight to an object just based on what we perceive visually, without having actually explored it haptically. It also gives rise to the assumption that other properties, such as object hardness, are dealt with similarly.

\subsection{Integration of Visual and Haptic Experiences}
\label{sec:bio:visuo-haptic}
The reliability of each sensory modality plays a crucial role in how our brain weighs and combines our visual and haptic experiences of an object to more abstract and meaningful concepts \citep{Helbig2007Optimal, Ernst2002Humans}. We are not born with this ability; it emerges and matures as we live and accumulate experiences of the world. While we do so, the neurons in our brain organize among themselves, a process which has been termed input-driven self-organization \citep{Miikkulainen2005Computational}.

The integration of multiple sensory modalities at the level of a single neuron has been studied in the cat superior colliculus \citep{Stein2014Development}. Newborn cats can already detect certain cross-modal correspondences, but the ability to integrate information from different senses develops after birth. The underlying neural circuitry adapts to the cross-modal experiences of the environment while optimizing the multisensory integration capabilities. This learning process does not wait for the contributing unisensory system to fully mature. Both the unisensory perceptual skills and the ability to integrate information from multiple senses develop in parallel.

A lot speaks for self-organization among the neurons being a fundamental principle for how the brain functions. One example is the neurons in the primary visual cortex that learn selectivity for certain features like orientation and colour and form different cortical feature maps \citep{Miikkulainen2005Computational}. The coarse structure of these feature maps is predetermined even before birth by retinotopy, while the more granular structure is shaped by visual experience after birth. The first few weeks seem especially critical: experiments have shown that depriving kittens of typical visual experience in this stage of their development can cause irreparable permanent physiological effects, even blindness \citep[e.g.,][]{Hubel1970Period, Blakemore1970Development, Blakemore1975Innate}.
The somatic sensory maps develop in a similar manner, possibly starting with the first body movements while still in the womb \citep{Mountcastle2005Sensory}.

A behavioural study performed by \cite{Gori2008Young} offers the most important evidence thus far on the role of input-driven self-organization in our acquiring of visuo-haptic integration capabilities. They found that a human's ability to integrate visual and haptic inputs related to object form becomes statistically optimal between the ages 8 and 10. The weight that children below that age range assign to either modality often does not correspond to their respective reliability in a particular situation. Further, perceptual illusions, such as the rubber hand illusion (RHI), indicate that the temporal co-occurrence between unimodal experiences is what triggers the creation of associative links between the sensory modalities \citep{Botvinick1998Rubber}. The likelihood of stimuli coming from the two modalities being integrated increases if it is known that they originate from the same object or are otherwise spatially related \citep{Helbig2007Optimal}.

\subsection{Organizational Principles}
\label{sec:bio:principles}

We do not have a complete picture of how object perception works in the brain and how visual and haptic cues are combined to accomplish object-related tasks. However, we can derive some basic principles from the evidence presented above that could help us build robots with human-like proficiency in object perception:

  \paragraph{Hierarchical processing} Object recognition and identification are performed by the ventral visual pathway, which starts in the occipital lobe and reaches down to the temporal lobe in the cerebral cortex. The processing of the visual input occurs in a hierarchical fashion along this pathway, with increasingly complex and abstract features being extracted. 
  \paragraph{Separate substreams for object shape and material perception} Some areas along the ventral pathway are responsive to haptic stimuli. Bimodal activation has been detected in the LOC, in charge of perceiving the geometric shape of objects. Neighbouring and sometimes crossmodally interacting foci specialized in the processing of material properties were identified in more medial areas of the ventral pathway, along with the CoS specifically. This evidence supports the idea that the ventral pathway is further organized into two substreams for object shape and material perception stretching across the more lateral and more medial areas, respectively.
  \paragraph{Input-driven self-organization} The ability to integrate the visual and haptic input in a statistically optimal way is not innate but emerges only after birth as we experience the world around us. Here, unimodal stimuli's temporal and spatial co-occurrence serves as a trigger for multimodal integration.

\section{Multimodal Object Perception in Robots}
\label{sec:sota-tech}
The previous section presented some organisational and functional principles that enable visuo-haptic object perception and recognition in the brain. The following sections cover the sensory and computational aspects used for visuo-haptic object perception and recognition in robots and other artificial systems and indicate how they relate to their biological counterpart. We start with a brief overview of visual sensors, follow up with the topics of tactile sensors, and continue with data collection and datasets.

\subsection{Visual Sensors}
Visual sensors or cameras are ubiquitous nowadays and designed to create images that are interpretable by humans. Although their working principle has been perfected in the past two hundred years \citep{Brady2018Parallel}, the field continues to evolve. However, due to the abundance of material for visual sensors and their applications, we will provide only a short overview of the most common technologies used in robotic applications before moving on to the less established tactile sensing technologies.

Cameras capturing visible light (400-700nm) have become commodities. Most of the research and application in robotics and computer vision have specialized in greyscale or RGB images obtained with these types of cameras. However, they have been optimized for human interpretation rather than computer vision and robotics. Moreover, their performance is significantly impacted by environmental conditions such as illumination intensity and direction, fog, haze, and smoke \citep{Gade2014Thermal}. Thus specialized solutions optimized for computation are needed. Some of these alternatives might be RGB-D, thermal cameras \citep{Gade2014Thermal}, parallel cameras \citep{Brady2018Parallel} or event cameras \citep{Gallego2022EventBased}.

Nowadays, some of the most common sensors used for visual perception in robotics are consumer-grade RGB or RGB-D cameras. RGB-D cameras provide a visible light (RGB) image and a depth image used for the 3D perception of a scene. These cameras produce depth images using near-infrared (NIR) light projection (750-1400nm) and different working principles, such as time-of-flight (ToF) for the Microsoft Kinect v2, structured-light (SL) for the Asus Xtion Pro Live, and active stereo vision (ASV) for the Intel Realsense R200 cameras \citep{Kuan2019Comparative}.

Thermal cameras capture infrared radiation. Although initially developed as a surveillance and night vision tool for the military, as the technology has matured and the price has dropped, their use has expanded to other fields of application such as robotics \citep{Gade2014Thermal}.

More recently, event cameras have also become popular in robotics research. They are bio-inspired sensors that asynchronously measure per-pixel changes and output a stream of events that encode the changes' time, location and sign. This operation principle translates to high temporal resolution, very high dynamic range, low power consumption, and high pixel bandwidth, which are attractive properties for mobile robotics, augmented and virtual reality (AR/VR), and video game applications \citep{Gallego2022EventBased}.

\subsection{Tactile Sensors}
\label{sec:sensors}
Tactile sensors are mostly designed to mimic mechanoreceptors, particularly to detect mechanical pressure. The main objectives of tactile sensors are to determine the location, shape and intensity of contacts. These properties are determined by measuring the instantaneous pressure or force applied to the sensor's surface on multiple contact points. Also, the contact's late effects, i.e., body-borne vibrations, may carry relevant information. Body-borne vibrations are not as commonly measured or exploited as part of haptic sensing; however, there are some examples, e.g., \cite{Syrymova2020VibroTactile, Toprak2018Evaluating}, including sensors that are inspired by hair follicle receptors or ciliary structure \citep{Alfadhel2015Magnetic, Ribeiro2017Bioinspired, Kamat2019Bioinspired} and that have been proven very effective in obtaining information about the texture of objects \citep{Ribeiro2020Highly, Ribeiro2020Fruit}.

Thermoceptors, although an integral part of human haptic perception, are typically not classified as tactile sensors within robotic applications. However, they are sometimes included because they might help compensate for thermal effects \citep{Tomo2016Design}, thus helping to obtain a more robust electronic signal related to pressure or vibrations, or because they might help to classify the material of the object in contact \citep{Wade2017Force}. In contrast, nociceptors have not yet been developed as part of haptic or tactile sensing per se but can be and have been implemented in software based on the limitations of robots \citep[e.g.,][]{Navarro-Guerrero2017Improving, Navarro-Guerrero2017Effects}. 

Technologies for tactile sensing have been developed since the early '70s and have greatly improved in the past ten years \citep{Dahiya2010Tactile, Dahiya2013Tactile, Kappassov2015Tactile}, but the field is still young, and there are no widely accepted solutions. Several transduction methods have been explored, including capacitive \cite[e.g.,][]{Larson2016Highly}, piezoelectric 
\cite[e.g.,][]{Seminara2013Piezoelectric}, piezoresistive \cite[e.g.,][]{Jung2015Piezoresistive}, optical \cite[e.g.,][]{Ward-Cherrier2018TacTip,Kuppuswamy2020SoftBubble}, fiber optics \cite[e.g.,][]{Polygerinos2010MRICompatible}, and magnetic \cite[e.g.,][]{Jamone2015Highly}. Table \ref{tab:transduction-mechanisms} summarizes the advantages and disadvantages of the different transduction principles for detecting mechanical pressure. For additional information, please refer to \cite{Chi2018Recent}.

\begin{table*}[!hbt]
  \centering
  \caption{Transduction mechanisms for detecting mechanical pressure. Tactile sensor design is benefiting from rapid nanomaterial and nanocomposite fabrication technology advancements. This table is based on \citet{Chi2018Recent}.}
  \label{tab:transduction-mechanisms}
  \begin{tabular}{@{}m{0.22\textwidth} m{0.24\textwidth} @{\hspace{1mm}}m{0.28\textwidth} m{0.19\textwidth}@{}}
  \hline
  \textbf{Transduction\newline Mechanisms} & \textbf{Advantages} & \textbf{Disadvantages} & \textbf{Example} \\ \hline \hline

  Capacitive \\[-12pt]
  \tablebodyfont
  based on the capacitance of parallel plates separated by an elastic dielectric layer.
   & 
    \begin{itemize}
    \tablebodyfont
    \item High spatial resolution
    \item High sensitivity
    \item Large dynamic range
    \item Temperature independent
    \end{itemize}
  & \begin{itemize}
    \tablebodyfont
    \item Stray capacitance
    \item Complex measurement circuit
    \item Cross-talk between elements
    \item Susceptible to noise
    \item Hysteresis
    \end{itemize}
  & \tablebodyfont 
  \citet{Larson2016Highly}
  \\[-5pt] \hline

  Piezoresistive \\[-10pt]
  \tablebodyfont
  based on the transduction of forces into resistance changes.
  & 
    \begin{itemize}
    \tablebodyfont
    \item High spatial resolution
    \item Low cost
    \item Simple construction
    \item Compatible with VLSI
    \end{itemize}
  & \begin{itemize}
    \tablebodyfont
    \item Hysteresis
    \item High power consumption
    \item Lack of reproducibility
    \end{itemize}
  & \tablebodyfont 
  \citet{Jung2015Piezoresistive}
  \\[-5pt] \hline

  Piezoelectric \\[-12pt]
  \tablebodyfont
  based on the transduction of forces into voltage changes.
  & 
    \begin{itemize}
    \tablebodyfont
    \item High sensitivity
    \item High dynamic range
    \item High frequency response
    \item High accuracy
    \end{itemize}
  & \begin{itemize}
    \tablebodyfont
    \item Poor spatial resolution
    \item Charge leakages
    \item Dynamic sensing only
    \item Temperature-dependent sensitivity and robustness
    \end{itemize}
  & \tablebodyfont 
  \citet{Seminara2013Piezoelectric}
  \\[-5pt] \hline

  Optical \\[-7pt]
  \tablebodyfont
  based on changes of light intensity modulation, interferometry or fiber Bragg grating.
  & 
    \begin{itemize}
    \tablebodyfont
    \item High spatial resolution
    \item Good reliability
    \item Wide sensing range
    \item High repeatability
    \end{itemize}
  & \begin{itemize}
    \tablebodyfont
    \item Non-conformable
    \item Bulky in size
    \item Susceptible to temperature or misalignment
    \end{itemize}
  & \tablebodyfont 
  \citet{Kuppuswamy2020SoftBubble, Ward-Cherrier2018TacTip}
  \\[-3pt] \hline

  Magnetic \\[-2pt]
  \tablebodyfont
  based on changes in the magnetic field caused by the mechanical deformation of a soft material.
  & 
    \begin{itemize}
    \tablebodyfont
    \item High sensitivity
    \item High dynamic range
    \item Linear output
    \item High power output
    \end{itemize}
  & \begin{itemize}
    \tablebodyfont
    \item Mechanical hysteresis (of the soft material)
    \item Affected by strong external magnetic fields
    \end{itemize}
  & \tablebodyfont 
  \citet{Jamone2015Highly,Paulino2017LowCost,Tomo2018New}
  \\[0pt] \hline
  \end{tabular}
\end{table*}

\subsubsection{Commercial Sensors}
Although there are some commercial solutions, the costs are still relatively high, and the performance level is not always satisfactory. In the remainder of this section, we present some of the commercial solutions for tactile sensing. Although we are aware of other commercial sensors, such as the WTS-FT by Weiss Robotics GmbH \& Co.\ KG., all but the presented here seem to have been discontinued at the time of writing.

\begin{figure}[htbp]
  \centering
  \includegraphics[height=0.09\textheight]{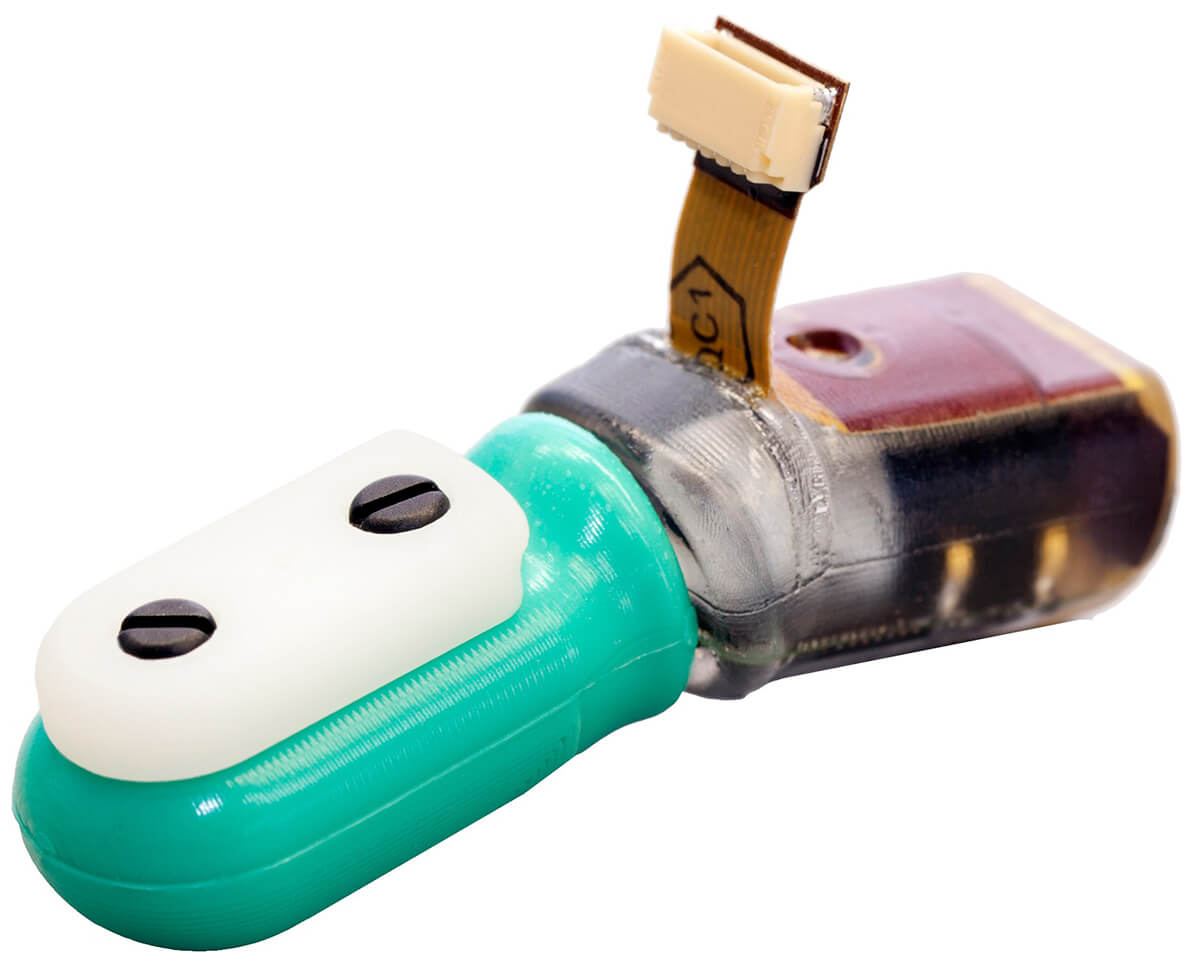}
  \includegraphics[height=0.08\textheight]{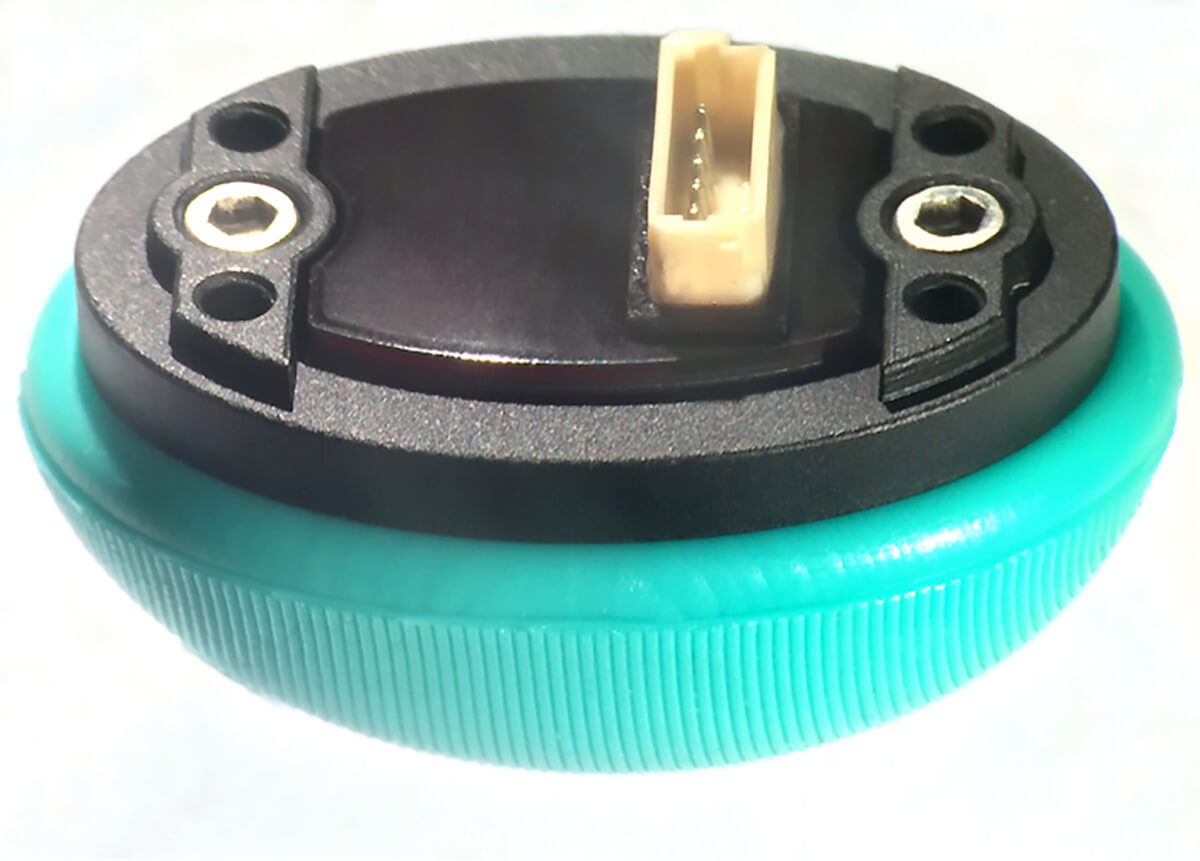}
  \includegraphics[height=0.08\textheight]{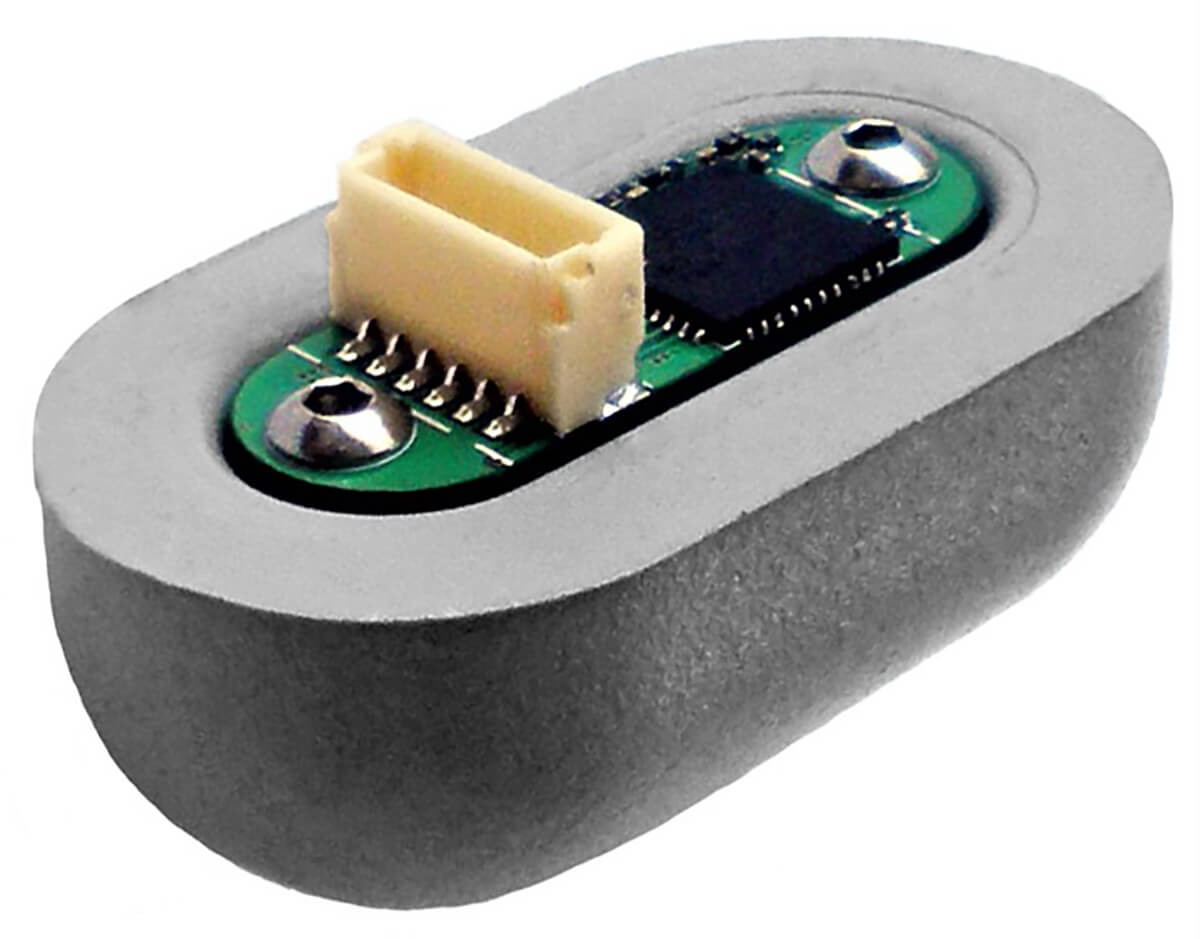}
  \caption{From the left: SynTouch\textsuperscript{\textregistered} BioTac\textsuperscript{\textregistered}, BioTac\textsuperscript{\textregistered} SP, and NumaTac\textsuperscript{\textregistered} Tactile Sensors. Images used with permission from SynTouch\textsuperscript{\textregistered} \url{https://syntouchinc.com/}.}
  \label{fig:SynTouch-biotac}
\end{figure}

The BioTac\textsuperscript{\textregistered} sensor by SynTouch\textsuperscript{\textregistered} was launched in 2008. The sensor's design attempts to mimic some of the human fingertip's physical properties and sensory capabilities. It consists of a rigid core surrounded by an elastic bladder filled with liquid. This construction provides a compliant surface, allowing it to sense force, vibration, and temperature. SynTouch\textsuperscript{\textregistered} offers variations of the technology tailored to different applications. Examples for robotic applications are shown in Figure \ref{fig:SynTouch-biotac}.

The DIGIT tactile sensor \citep{Lambeta2020DIGIT} by GelSight is an optical tactile sensor using a piece of elastomeric gel with a reflective membrane coat on top, which enables it to capture fine geometrical textures as a deformation in the gel. A series of LEDs with RGB colour illuminates the gel such that a camera can record the deformation.

\begin{figure}[htbp]
  \centering
  \includegraphics[height=0.18\textheight]{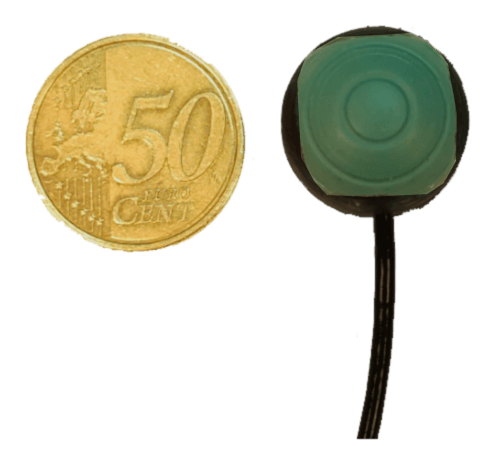}
  \includegraphics[height=0.18\textheight]{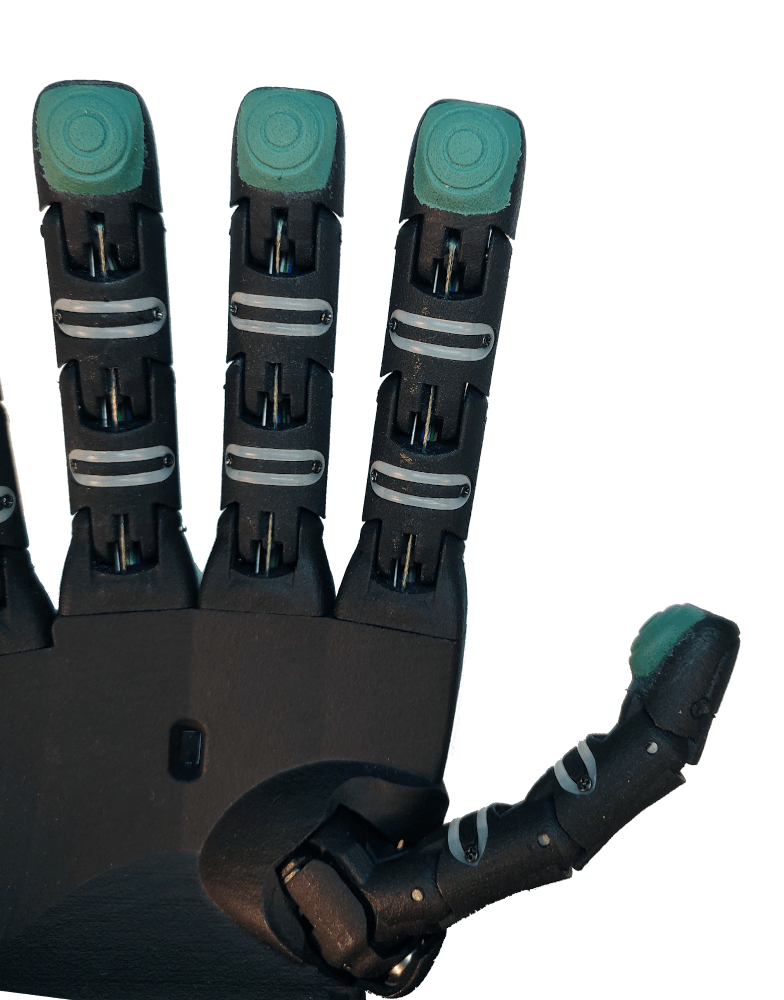}
  \caption{Left: the SINGLEX stand-alone tactile pressure sensor version. Right: FTS tactile pressure sensor mounted on a robot finger. Images used with permission from Seed Robotics \url{https://www.seedrobotics.com/}.}
  \label{fig:seedrobotics}
\end{figure}

Seed Robotics' FTS Tactile pressure sensors (see Figure \ref{fig:seedrobotics}) are low-cost sensors that offer high-resolution contact force measurement (1mN/0.1g resolution up to 30N range). The sensor compensates for temperature, and it is immune to magnetic interference. The sensors are directly integrated into the robotic hands also offered by the company. However, there is a stand-alone version of the sensor for use in third-party user applications.

\begin{figure}[htbp]
  \centering
  \includegraphics[height=0.105\textheight]{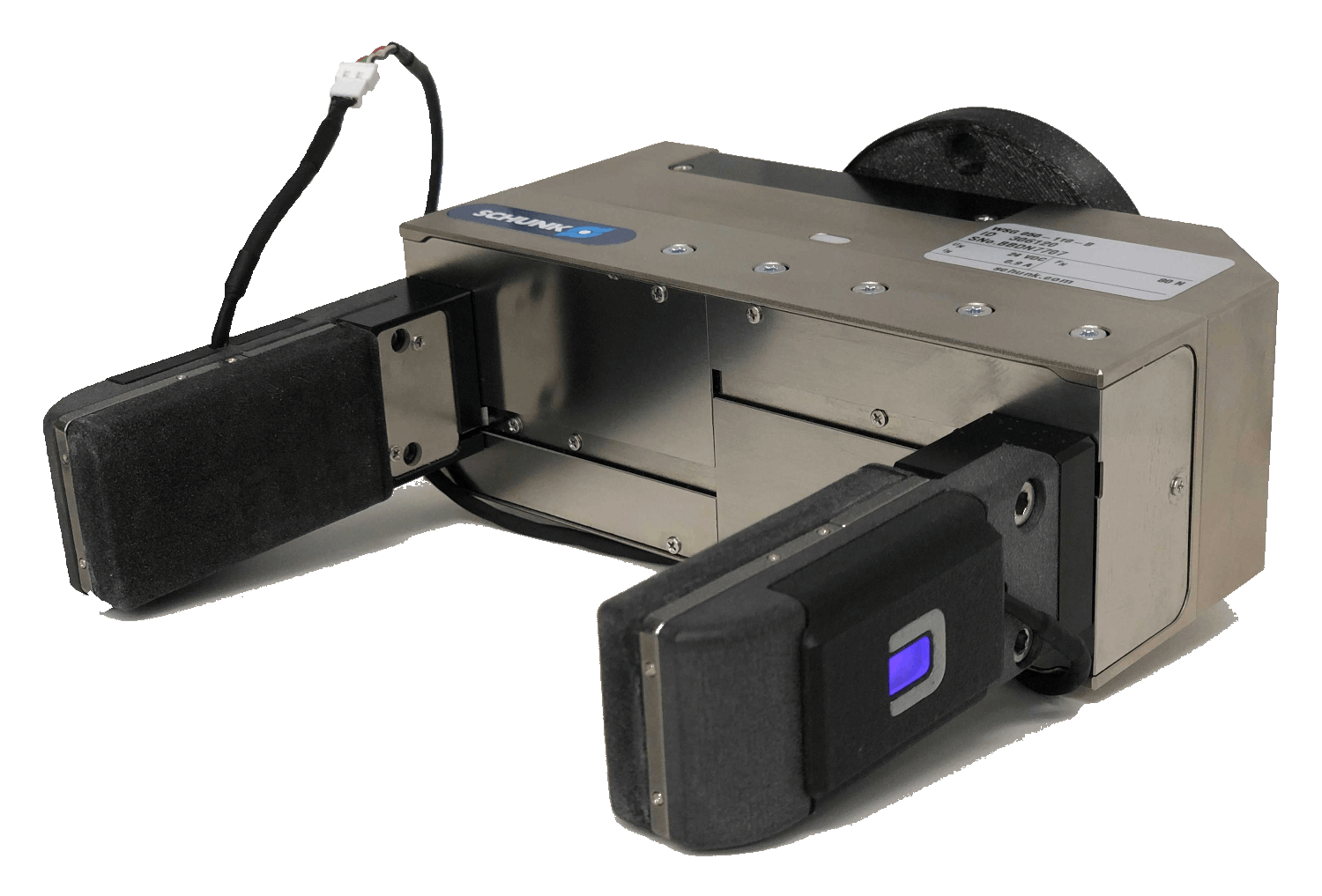} \hfill
  \includegraphics[height=0.105\textheight]{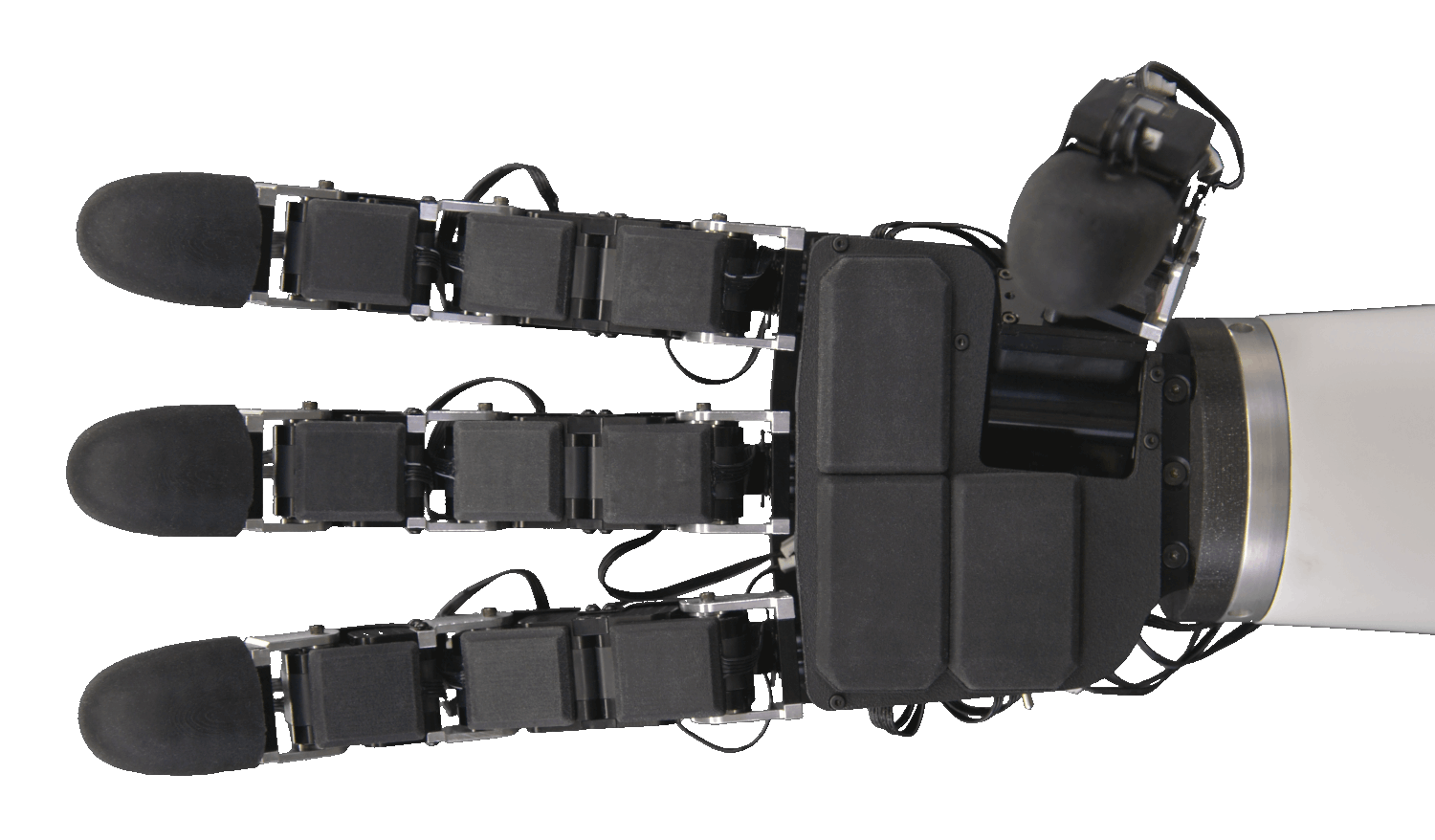}
  \caption{Left: a flat version inspired by \cite{Tomo2018New}. Right: a curved version inspired by \cite{Tomo2018Covering}. Images with permission from Xela Robotics \url{https://xelarobotics.com/}.}
  \label{fig:uskin}
\end{figure}

The uSkin sensor by Xela Robotics is a magnetic tactile sensor composed of small magnets embedded in a thin layer of flexible rubber and placed above a matrix of magnetic Hall-effect sensor chips. 
Upon contact, the magnets are displaced and the magnetic field sensed by the Hall-effect chips changes; the contact forces can be estimated from these variations in the magnetic field. 
The uSkin sensor can measure the full 3D force vector (i.e., both normal and shear contact forces) at each tactel, with a good spatial resolution (about 1.6 tactels for square cm), high sensitivity (minimum detectable force of 1gf), and high frequency ($>100$Hz, depending on the configuration). 
Different versions of the sensor are available to cover both flat and curved surfaces, see Figure \ref{fig:uskin} for an example.

\begin{figure}[htbp]
  \centering
  \includegraphics[height=0.12\textheight]{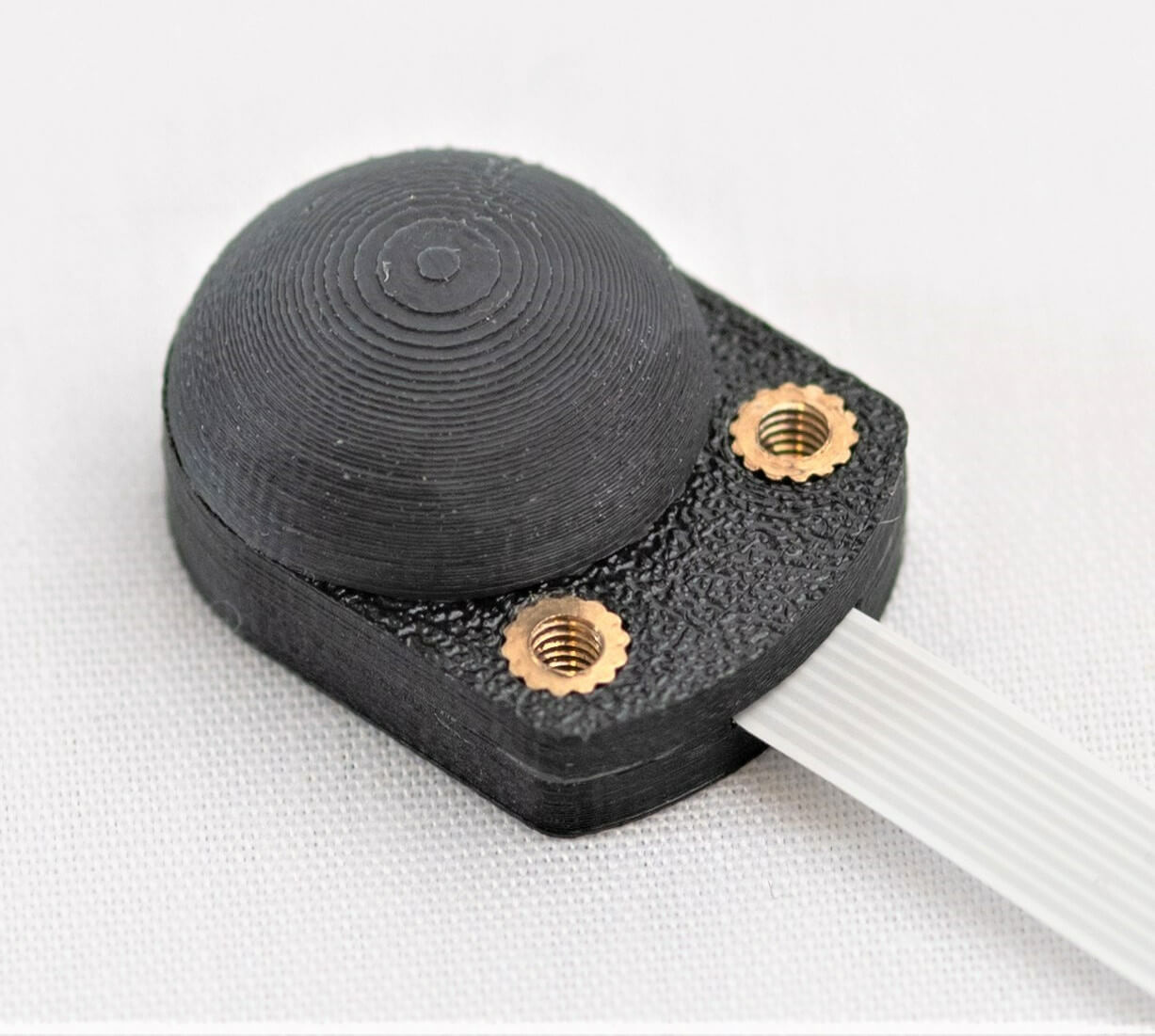}
  \includegraphics[height=0.12\textheight]{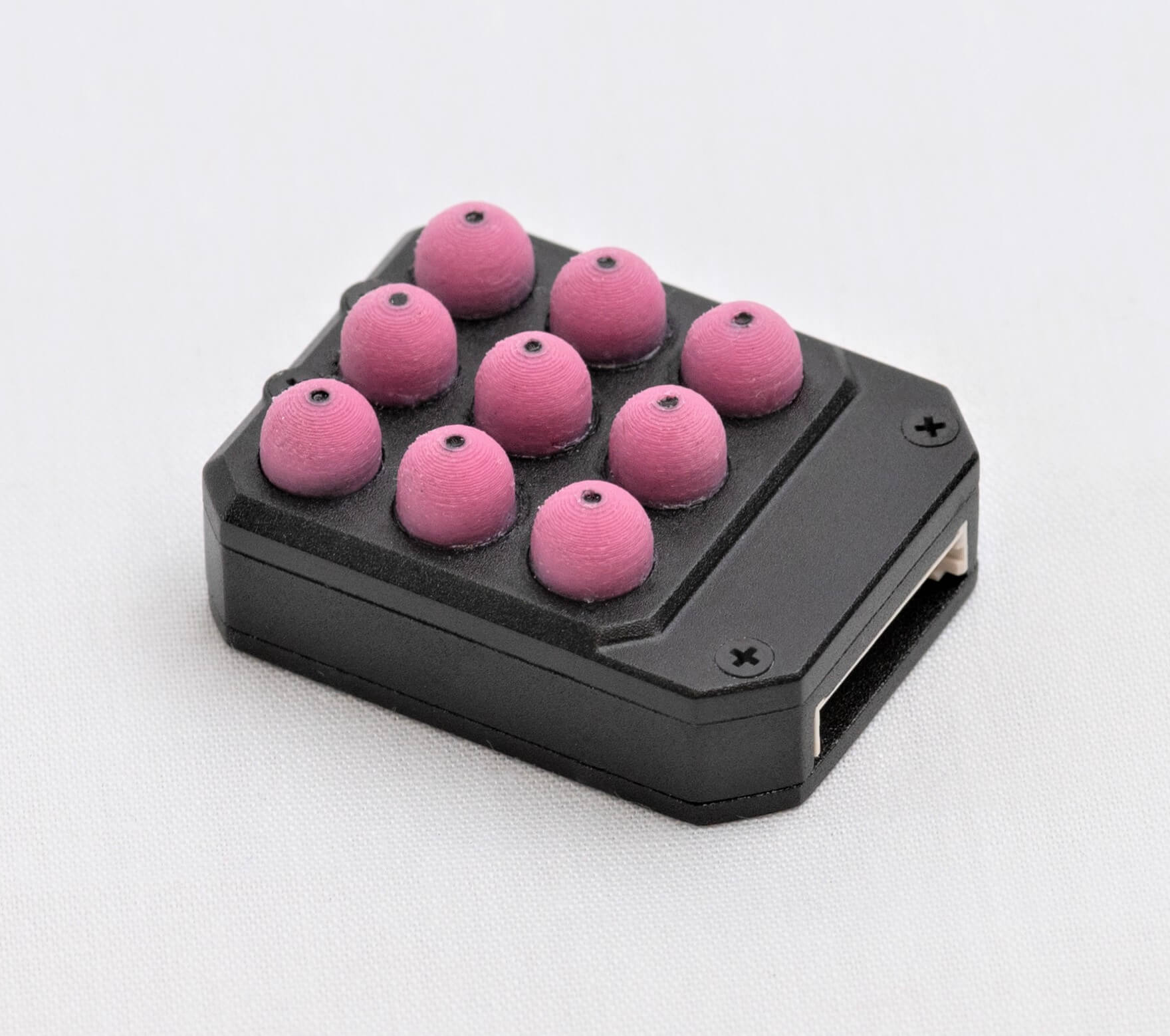}
  \caption{Left: Single 3D force tactile sensor. Right: A slim tactile sensor array (PapillArray Sensor) available in different configurations. Images from Contactile \url{https://contactile.com/} licensed under CC BY-NC-ND 4.0.}
  \label{fig:contactile}
\end{figure}

Finally, Contactile offers both a stand-alone sensor and tactile sensor arrays called PapillArray sensor, see Figure \ref{fig:contactile}. These optical sensors consist of infrared LEDs, a diffuser, and four photodiodes encapsulated in a soft silicone membrane. The photodiodes are used to measure the light intensity patterns to infer the displacement and force applied to the membrane. This strategy allows for the measurement of 3D deflections, 3D forces and 3D vibrations, as well as the inference of emergent properties such as torque, incipient slip and friction.

The need for such technologies is pushing research forward in the development of both, new sensing technologies and applications such as robotic grasping, smart prostheses, and surgical robots. 
In particular, enhancements are still needed in a number of aspects (e.g., mechanical robustness, sensitivity and reliability of the measurements, ease of electromechanical integration and replacement) to deploy sensors in practical applications.

Of particular interest are solutions that: are flexible \citep{Larson2016Highly, SenthilKumar2019Review}, stretchable \citep{Bhattacharjee2013Tactile,Buescher2015Flexible} and can cover sizeable \citep{Dahiya2013Directions} and multi-curved \citep{JuinaQuilachamin2023Fingerprint, Tomo2018Covering} surfaces (possibly with a small number of electrical connections \citep{JuinaQuilachamin2023Fingerprint}), can detect multiple contacts at the same time \citep{Hellebrekers2020Soft}, can detect both normal and shear forces \citep{Tomo2018New}, can dynamically change the range and sensitivity of the measurements depending on the task \citep{holgado2018adjustable}, are affordable and can be easily manufactured \citep{JuinaQuilachamin2023Fingerprint, Paulino2017LowCost}. 
For more information on experimental tactile sensing technologies see \citet{Chi2018Recent}, and for a specialized review of printable, flexible and stretchable tactile sensors, see \citet{SenthilKumar2019Review}.

\subsection{Data Collection and Datasets}
\label{sec:datasets}
Data acquisition from tactile sensors still lacks a unified theoretical framework. Besides the sensor itself, tactile data is affected by the sequence of exploration procedures (EPs, see Section \ref{sec:bio:haptics}) and the application in which it is to be used in, among others. A single grasp can only perceive a portion of an object's properties, and the perception is limited to the surface that comes in contact with the tactile sensors. Thus, it is difficult, if not impossible, to recognize all properties of an object using one single tactile EP. Unlike vision, tactile perception is intrinsically sequential.

Authors such as \cite{Kappassov2015Tactile}, and \cite{Liu2017Recent} have defined tactile object recognition into subcategories in an attempt to create a unified framework for data collection. \cite{Kappassov2015Tactile} propose to divide tactile perception into tactile object identification, texture recognition, and contact pattern recognition. Whereas \cite{Liu2017Recent} propose to divide tactile perception into perception for shape, perception for texture, and perception for deformable objects. However, there is still no consensus on how to collect and organize data for haptic or visuo-haptic object recognition datasets.

In this section, we provide examples of datasets for multimodal object recognition and grasping.

\subsubsection{Datasets for Multimodal Object Recognition}
\label{sec:datasets-recognition}
One example of such a dataset comes from \cite{Kroemer2011Learning}, who generated a small-scale multimodal dataset for dynamic tactile sensing. Tactile information was collected using a custom whisker-like tactile sensor whose data resembles the \textit{Lateral Motion} EP. Data were collected for a total of 26 surfaces of 17 different materials. Visual information was collected by taking four grayscale pictures of those objects from different perspectives.

\cite{Sinapov2014Grounding} created a multimodal object recognition dataset comprising proprioceptive, auditory, and visual information but not tactile information. The dataset consists of 100 objects from 20 different categories. All objects were explored five times, using nine haptic interactions, and photographed. 
The interactions were not extensively described and thus cannot be confidently mapped to Lederman's EPs. They included press and poke (\textit{Pressure}), grasp (\textit{Enclosure}), lift, hold and push (app.\ \textit{Unsupported Holding}), plus tap, drop and shake, which seems to be primarily related to gathering auditory information, as well as the corresponding RGB image of the objects or an RGB video while performing the EPs.

\cite{Chu2015Robotic} collected a small-scale multimodal dataset for haptic perception, known as the Penn Haptic Adjective Corpus 2 (PHAC-2). The PHAC-2 dataset consists of haptic data collected with a pair of SynTouch\textsuperscript{\textregistered} BioTac\textsuperscript{\textregistered} sensors, which were mounted on the grippers of a Willow Garage PR2 robot. The labels were collected in a human study, where 25 haptic adjectives were assigned to the objects. 
The PHAC-2 dataset contains haptic and visual data for 60 household objects. Given the robot's and BioTac\textsuperscript{\textregistered} sensors' physical constraints, the objects were chosen to fit the following physical characteristics: the objects had to be between 15 and 80mm in width and a minimum height of 100mm. There were no restrictions regarding weight since the objects were not lifted. All objects included needed to be at room temperature, clean, dry, and durable. Furthermore, the object could not be sharp or pointed. 
Haptic data were collected for four EPs, namely, \textit{Pressure} (Squeeze), \textit{Enclosure} and \textit{Static Contact} (Hold), \textit{Lateral Motion}. The dataset includes two versions of the \textit{Lateral Motion} EP. The first version, referred to as \textit{slow slide}, is performed with low velocity and substantial contact force, and the second version, called the \textit{fast slide}, is of higher speed and half the contact force as for a slow slide. Every EP was repeated ten times per object, and the objects were re-positioned each time. Meanwhile, the visual data consists of high-resolution images of each object from eight different viewpoints.

Another small-scale dataset for visuo-haptic object recognition comes from \cite{Toprak2018Evaluating}. A NAO robot (model T14: torso-only) was used. Visual data was collected using one of the two RGB cameras in NAO's head. For the kinesthetic properties, namely, global shape and weight, the joint angles and the electric currents in the motors in both arms were measured when performing the respective EPs. 
For texture and hardness, inexpensive contact microphones were attached as sensors to NAO's arm and a custom-made table, on which it performed the corresponding EPs to capture the resulting vibrations transmitted across the surfaces. 
A total of 11 everyday objects were carefully selected to cover both visually and haptically ambiguous objects. Of each object, ten observations were collected under optimal lighting conditions (controlled and reproducible lab conditions) and another three under real-world lighting.

More recently, \cite{Bonner2021AU} created a public dataset for visuo-haptic object recognition containing information of 63 different objects. The visual information comes from high-resolution RGB images collected using near-ideal lighting conditions. 
The kinesthetic data was collected with the RH8D Robotic Hand by Seed Robotics using the \textit{Unsupported Holding} and \textit{Enclosure} EPs. The tactile information was captured using contact microphones mounted on the RH8D hand and on a NAO robot that was used to perform the \textit{Lateral Motion} and \textit{Pressure} EPs.

\subsubsection{Datasets for Multimodal Object Perception for Manipulation}
\label{sec:datasets-grasping}
\cite{Calandra2017feeling} provided a dataset for evaluating grasp success. Their hardware setup consisted of a 7-DoF Sawyer manipulator equipped with a WSG-50 gripper, one GelSight tactile sensor for each of the two gripper fingers and a Kinect V2 camera placed in front of the robot. First, using the Kinect's depth information, the object's position on a table in front of the robot was inferred. The gripper was randomly positioned above the object with its fingers opened. Next, a closing action was executed, and the gripper was lifted from the table.
After the lifting action, the tactile and visual information was used to infer whether the object was still on the table or successfully grasped. A label indicating the grasp success was automatically generated. The dataset collected through this automated data collection procedure consists of a total of 9269 grasp samples for 106 different objects. 

Another visuo-tactile dataset for grasping and related tasks such as slip-detection or visuo-tactile object classification is presented by \cite{Wang2019multimodal}. They used two Intel RealSense SR300 cameras and a UR5 robot arm equipped with an Eagle Shoal hand with piezoresistive tactile sensors. The objects to be grasped were 10 everyday grocery items like detergent bottles or soup cans, intentionally selected to be container-like and either full or empty for generating different tactile readings. The dataset includes 2550 grasping attempts containing information like RGB and depth images from different grasp stages and videos of the whole grasp, tactile information from the 16 tactile sensors included in the hand and ground truth information including timestamps and grasp outcome. 

In the same direction, \cite{Li2018Slip} introduced a dataset for slip detection during manipulation. Their setup consisted of a 6-DoF UR5 robot arm and a WSG-50 parallel gripper, with one gripper's finger replaced by a GelSight sensor for tactile recordings, and a regular webcam mounted on the side of the gripper for visual recordings. The authors thresholded the relative displacement between the object's texture and markers of the GelSight sensor during a grasp attempt to detect if a slip occurred. The dataset covers examples of translational, rotational and incipient slips. The data acquisition was done by taking a sequence of consecutive tactile and corresponding camera image pairs at a frequency of 20 Hz. Their dataset consists of 1102 grasp-and-lift attempts on 84 different household objects with varying sizes, shapes, surface textures, materials and weights. The authors provide data of 152 grasp attempts on 10 additional objects for testing purposes.

While the previously presented datasets use a robot to collect the information, some datasets of human grasping can also be used to train robotic grasping. For instance, \cite{Brahmbhatt2019Contactdb} provide a multimodal dataset from human grasps of household objects. Participants were instructed to grasp 3D-printed objects with a specific post-grasp functional intent. Different post-grasp functional intents lead to different grasping approaches, even for the same object, e.g., when instructed to hand it off vs to use it. The contact surface of the hand with the object represents the haptic modality, which is captured by a FLIR Boson 640 thermal camera. In contrast, the visual modality is represented by RGB-D images collected with a Kinect V2 camera. The dataset contains 375 000 synchronized RGB-D and thermal images collected during grasping 50 different household objects, giving rich information about human grasps through detailed contact maps.

\section{Multimodal Machine Learning}
\label{sec:methods}
Once the multimodal data from the sensors, such as those presented in Section \ref{sec:sensors}, has been collected, it needs to be processed and integrated to make it useful. 
Relying on different sensory modalities offers several advantages, as discussed in Section \ref{sec:bio:visuo-haptic}. However, the heterogeneity of the data (cf.\ Section \ref{sec:datasets}) creates multiple challenges. Understanding these challenges can help in applications and guide the development of new signal processing methodologies to deal with the complexities of multimodal information. In particular, \citet{Baltrusaitis2019Multimodal} identifies five core challenges: representation, translation, alignment, fusion and co-learning. 

In the rest of this section, we outline these general challenges and comment on how they relate to the concrete case of visuo-haptic perception in robotics to facilitate the understanding of architectural decisions and design choices for approaches presented in Section \ref{sec:applications}.

\subsection{Representation} 
\label{sec:mml-representation}
The first challenge refers to creating or learning a meaningful representation that allows for the preservation and exploitation of the complementarity or redundancy of the multiple modalities. 
A representation or feature vector/tensor can be an image, an audio sample, or discrete values such as \textit{open} or \textit{close}. 
Some of the challenges in creating useful representations from multimodal data are: 
\begin{itemize}
  \item how to deal with different levels of noise?
  \item how to deal with missing data?
  \item how to deal with out of phase signals or different frequency rates?
  \item how to deal with different vector sizes?
\end{itemize}

\citet{Bengio2013Representation} suggested some desirable properties for representations, including: 
\begin{description}
  \item[\textit{Smoothness}:] similarity of concepts should be preserved in the representation space. 
  \item[\textit{Natural clustering}:] different concepts should lead to differentiated representations.
  \item[\textit{Temporal and spatial coherence}:] consecutive (for sequential data) or spatially close observations should be associated with relevant regions of the representation space.
  \item[\textit{Sparsity}:] most extracted features should be insensitive to minor variations of any given observation.
  \item[\textit{Expressive}:] should capture a large number of possible input configurations.
  \item[\textit{Distributed}:] to allow for reuse and recombination of the activation of parameters or subsets of features across concepts.
  \item[\textit{A hierarchical organization of explanatory factors}:] increasingly abstract features should be defined in terms of less abstract ones.
\end{description}

More recently, \citet{Baltrusaitis2018Challenges, Baltrusaitis2019Multimodal} proposed two categories of multimodal representation: \textit{joint} and \textit{coordinated} representations. 
\textit{Joint representations} take all the available modalities as input and are used to create a single joint representation. 
In \textit{coordinated representations} each modality is used to create an independent representation. However, intermediate features across modalities are `coordinated' using similarity or structure constraints. Similarity-based coordination could, for instance, minimize a distance metric between the features. 
In structure-constrained coordination, constraints such as order are used. Examples of structure-constrained coordination are hashing, cross-modal retrieval, and image captioning \citep{Baltrusaitis2018Challenges, Baltrusaitis2019Multimodal}.

The modality representation also affects the fusion strategy (see Section \ref{sec:mml-fusion}), e.g., while optical tactile sensors such as GelSight or vibration data via spectrograms could allow for early integration with visual data, kinesthetic information would likely not.

\subsection{Translation / Mapping}
\label{sec:mml-mapping}
A second challenge concerns the translation or mapping of data from one modality to another. In addition to the heterogeneity of the data, the mapping is often not unique and potentially subjective. Thus, the evaluation of the mapping becomes a challenge \citep{Baltrusaitis2019Multimodal, Baltrusaitis2018Challenges}. 

\citet{Baltrusaitis2019Multimodal, Baltrusaitis2018Challenges} indicate that several machine learning applications correspond to translation between two modalities, such as automated text translation, image or video captioning, and speech transcription. In the context of multimodal object perception, translation could, for instance, serve as a mechanism to deal with the absence of a modality. 

\citet{Baltrusaitis2019Multimodal} further categorize multimodal translation into two categories: \textit{example-based} and \textit{generative}. 
Example-based models use a dictionary, which makes models large, task-specific and unwieldy. 
In contrast, generative approaches construct a model to perform the translation. However, generative models are challenging to build as they require understanding both the source and target modality \citep{Baltrusaitis2019Multimodal}. 

Three broad categories can be identified within generative models: \textit{rule-based}, \textit{encoder-decoder}, and \textit{continuous generation} models \citep{Baltrusaitis2019Multimodal}. 
Rule-based models rely on pre-defined rules to translate features. They are more likely to generate syntactically or logically correct translations. Typically, the representation of each modality should share similarities with the representations of the other modalities; for example, \cite{Falco2017CrossModal} employ point clouds as a visuo-haptic common representation, and they combine data pre-processing, feature engineering and transfer learning techniques to realize an effective mapping.
In fact, this category of approaches often requires complex pre-processing pipelines to create the features used for the translation \citep{Baltrusaitis2019Multimodal}.
Encoder-decoder models, on the other hand, encode the source modality to a latent representation which is then used by a decoder to generate the target modality \citep{Keren2018Deep}, reducing the requirements of data pre-processing and feature engineering, although typically requiring larger amounts of data to obtain effective mappings.

Continuous generation models generate the target modality continuously based on a stream of source modality inputs and are most suited for translating between temporal sequences. In general, these models require temporal consistency between modalities \citep{Baltrusaitis2019Multimodal}; however, learning from weakly-paired training data has been recently attempted by \cite{Liu2019Active}, using sparse dictionary learning.

\subsection{Alignment}
\label{sec:mml-alignment}
Determining the relationship between features across modalities is another challenge for multimodal machine learning \citep{Baltrusaitis2019Multimodal, Baltrusaitis2018Challenges}. 
Similarly, as for the \textit{translation} challenge, here, the evaluation metrics might be the primary challenge. However, other challenges exist, such as the availability of datasets for evaluation, long-range dependencies and ambiguities, and the lack of correspondence between modalities. 

\citet{Baltrusaitis2019Multimodal} identifies two types of alignment: \textit{explicit} and \textit{implicit}. For explicit alignment, the alignment is obvious and easier to measure, such as in automatic video captioning or in the context of visuo-haptics, the alignment between thermal and RGB-D images in the multimodal dataset of \cite{Brahmbhatt2019Contactdb} presented in Section \ref{sec:datasets-grasping}. While for implicit alignment, a latent or intermediate representation is used, for instance, image retrieval based on text description where words are associated with regions of an image, or visuo-tactile fusion learning methods with self-attention mechanisms \citep{Cui2020SelfAttention}.

Aligning features across modalities could be necessary to exploit the complementarity of the different modalities.

\subsection{Fusion}
\label{sec:mml-fusion}
A fourth challenge is to join information from multiple modalities. 
Three approaches can be identified based on how the information from different modalities is combined: \textit{pre-mapping}, \textit{midst-mapping} and \textit{post-mapping} fusion \citep{Sanderson2004Identity, Toprak2018Evaluating}. These strategies are also referred to as \textit{early}, \textit{intermediate}, and \textit{late} integration \citep[e.g.,][]{Keren2018Deep}.

\begin{figure}[htbp]
  \centering
  \includegraphics[width=\columnwidth]{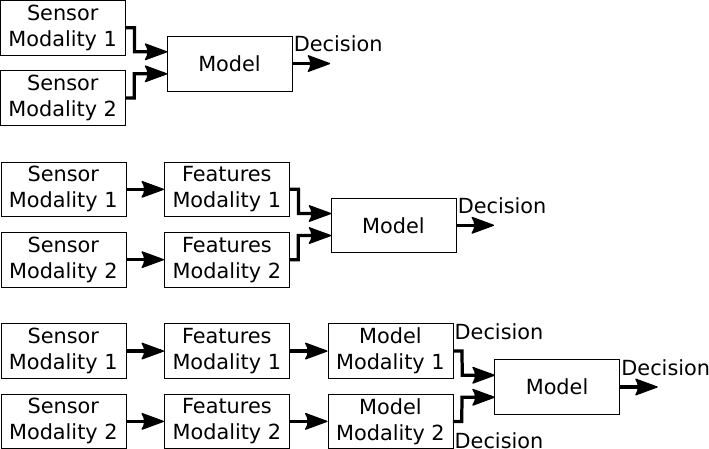}
  \caption{Information Fusion Strategies. Example with two modalities. Top: Monolithic or \textit{pre-mapping} fusion. Middle: \textit{midst-mapping} fusion. Bottom: \textit{post-mapping} fusion, the feature modalities modules are not strictly necessary.}
  \label{fig:fusion-strategies}
\end{figure}

In \textit{pre-mapping} fusion, the feature descriptors from the different modalities are concatenated into a single vector prior to the mapping into the decision space. While this strategy is simplistic and hence easy to implement, the disadvantage is that each modality's impact on the result is fixed as it depends on the respective feature vector's size instead of its statistical relevance. 
In \textit{midst-mapping} fusion, the feature descriptors are provided to the model separately. The model then processes these descriptors in separate streams and integrates them while performing the mapping. 
Lastly, in \textit{post-mapping} fusion, each feature descriptor is first mapped into the decision space separately, after which the decisions are combined to a final result. Figure \ref{fig:fusion-strategies} illustrates the different information strategies. 

Apart from being the most frequently used, \textit{midst-mapping} fusion appears to be the most promising among these three approaches as far as performance is concerned \citep{Castellini2011Using}. Moreover, this integration strategy would also be the best choice considering the principles on how multimodal object recognition is organized in the brain, as outlined in Section \ref{sec:bio:principles}, since the hierarchical processing in substreams that later converge to a decision can be modelled with it. This kind of setup has been used extensively with two substreams focusing on processing visual and haptic inputs separately. Nevertheless, to the best of our knowledge, only \cite{Toprak2018Evaluating} have investigated all three principles simultaneously, also including the separate processing of \textit{object shape} and \textit{material properties} in two substreams as well as the use of self-organizing mechanisms for processing and integration of the information.

\subsection{Co-learning or Transfer Learning} 
\label{sec:mml-co-learning}
The final challenge described by \citet{Baltrusaitis2018Challenges} is co-learning. 
Co-learning is described as a more general form of transfer learning at the level of representation or inference. Co-learning is particularly useful when data for some modality is limited, and information from a different modality can be used to aid training by exploiting complementary information across modalities. Thus, it is particularly relevant in multimodal object perception, where visual data is ubiquitous and tactile data is scarce. Co-learning is task-independent and could be used in fusion, translation, and alignment models \citep{Baltrusaitis2018Challenges}. 

\citet{Baltrusaitis2019Multimodal} identified three types of co-learning approaches: \textit{parallel}, \textit{non-parallel}, and \textit{hybrid}. Parallel-data approaches required observations from the same dataset and instances. In contrast, non-parallel data approaches can use data from a different dataset with overlapping classification categories. Finally, hybrid data approaches use a shared modality or dataset to achieve the transfer \citep{Baltrusaitis2019Multimodal}.
More recently, \citet{Rahate2022Multimodal} further extended this taxonomy to include cases for missing modalities, the presence of noise, annotations, domain adaptation, and interpretability and fairness. For a complete description of the taxonomy and examples, please see \citet{Rahate2022Multimodal}.

The reduced number and small size of public datasets for multimodal object perception motivates the study of transfer learning from visual object recognition to tactile object recognition. Such initiatives would also help to cope with the diverse number of robot embodiments, i.e., different sensors and actuators, which hinders progress on multimodal object perception. However, knowledge transfer from one modality to another is still an incipient field of research.

\section{Applications of Multimodal Object Perception}
\label{sec:applications}
This section presents examples of multimodal object perception applications, from object recognition, peripersonal space representation, and object manipulation. However, due to the heterogeneity of the applications, experimental setups and datasets, no cross-comparison will be provided. Hence, some examples are shown to provide a glance into the state of the art of multimodal object perception applications.

\subsection{Multimodal Object Recognition}
\label{sec:recognition}
Object recognition and the recognition of their properties are crucial for effective interaction with them both in biological and artificial systems. As such, an extensive body of work in this field exists. Here, we provide an overview of the techniques commonly used to address this problem.

\subsubsection{Unsupervised Learning}
\citet{Toprak2018Evaluating} presented a brain-inspired architecture for visuo-haptic object recognition, as outlined in Section \ref{sec:bio:principles}. Toprak et al.\ implemented an architecture including main principles identified in the processing of object-related stimuli in the brain, which are 1) hierarchical processing, 2) the processing of stimuli separated by object properties rather than by modality, and 3) experience-driven learning. Toprak et al.\ compared their brain-inspired architecture against a monolithic architecture or \textit{pre-mapping} fusion, where the features of all modalities were concatenated before processing, and a modality-based integration strategy, where visual and haptic features were preprocessed in two separate streams before being integrated into a final object classifier. Both of these strategies are commonly used in multimodal learning. To explore whether the brain-inspired processing principles could be useful for artificial agents, Toprak et al.\ implemented all three processing architectures using growing when required (GWR) neural networks on the same dataset and preprocessed input vectors. The hyperparameters for each architecture were optimized separately using hyperopt. The results indicate that hierarchical processing was indeed beneficial. 
However, results for the other two principles were not conclusive, and further research is needed. 
Toprak et al.\ further indicated that the size and quality of the dataset used might have played an essential role in exploring the value of processing object properties versus modalities in different streams.

\subsubsection{Supervised Learning}
\cite{Gueler2014What} used \textit{pre-mapping} fusion to classify the content of containers. The containers were squeezed, and both pressure and perceived visual deformation were used for classification. A three-fingered Schunk Dextrous Hand with pressure-sensitive tactile sensors was used to collect the haptic information, and an RGB-D camera placed 1 meter away was used to collect the visual data, but only a small region of interest around the finger of the robots was used for classification. The Tetra Pak containers were either empty or filled 90\% with water, yoghurt, flour, or rice. The collected data from multiple grasps was classified using k-means, quadratic discriminant analysis (QDA), k-nearest neighbours (kNN), and support vector machines (SVM). The results show that either modality is sufficient to perform the classification in this case, but classification accuracy can improve up to around 3\% under the tested conditions when the modalities are combined.

\cite{Corradi2017Object} compared one \textit{pre-mapping} fusion approach and two \textit{midst-mapping} fusion approaches. They used an optical tactile sensor, which consists of an illuminated ballon-like silicone membrane, and an internal camera detecting the shadow patterns created on the membrane. The camera images were processed using Zernike moments, which provided rotational invariance, and then PCA was used for dimensionality reduction. The visual data was processed using a bag-of-words (BoW) model of SURF features. The visuo-tactile recognition process was then performed in three manners: (1) for the \textit{pre-mapping} fusion approach, the unimodal feature vectors were concatenated, and kNN was used for classification, for the \textit{midst-mapping} fusion approaches, the posterior probabilities (the probability of the label given the observation) were estimated for each modality, and the classification was performed based on (2) either on the object label that maximizes their product or (3) the object label that maximizes the sum of these posterior probabilities weighted by the number of training samples available for each modality. Corradi et al.\ showed that multimodal classification achieves higher classification accuracy than either modality alone, and the posterior product approach achieves the highest classification accuracy among the tested approaches.

\cite{Bhattacharjee2018Multimodal} combined haptic information (i.e., force + motion) with thermal sensing to recognise objects in daily living environments. Several machine learning techniques were compared to train and test classifiers on a dataset of more than 60 objects. The data were collected with different robot movements (e.g., speed, direction) and at different times of the day (e.g. morning, afternoon, night) to reproduce the variability encountered in real-world conditions, generating significant differences in the haptic and thermal information. The results highlighted the importance of using multimodal information, especially in very unstructured environments characterised by high variability of the sensing conditions.

\cite{Liu2017VisualTactile, Liu2018Visual} implemented a \textit{midst-mapping} fusion approach using a kernel sparse coding method. Liu et al.\ used a three-fingered BarrettHand with capacitive tactile sensors in all three fingers and the palm. The tactile sensors have 24 taxels per finger with a spatial resolution of 5mm. The tactile information was processed using the canonical time-warping (CTW) method. At the same time, they used the covariance descriptor to characterize the visual information. The dataset consisted of 18 household objects. In general, kernel sparse coding (KSC) uses the idea that a signal can be reconstructed as a linear combination of atoms from a dictionary with which the data can then be encoded sparsely. However, this method fails to capture the intrinsic relations between the different data sources, and thus it can only be applied to each modality separately. To address that problem, Liu et al.\ proposed the joint group kernel sparse coding (JGKSC). Their results showed that fusing the visual and tactile information using the JKGSC method led to a higher classification accuracy than applying kNN or KSC to each modality separately.

More recently, deep learning methods have also started to be used in multimodal object recognition. For instance, \cite{Gao2016Deep} implemented a deep learning-based \textit{midst-mapping} fusion approach and tested it on the PHAC-2 dataset. 
The haptic data from the two BioTac\textsuperscript{\textregistered} sensors were normalized and downsampled to match the lowest sampling rate. Four out of 19 of the electrode impedance channels were selected using PCA. 
Data augmentation of the data was performed in two ways. Firstly, the two sensor readings were used as two distinct instances. Secondly, when downsampling the data, five different starting points were selected. Gao et al.\ suggest that the signal from both sensors and different downsampling strategies was highly similar, which resulted in overfitting of the CNN model used. 
The visual CNN model was based on the GoogleNet architecture pre-trained on the Materials in Context Database (MINC). 
The preprocessing of the visual data consisted of subtracting the mean values from the RGB image and resizing it using a central crop.
Finally, both feature vectors resulting from the haptic and visual networks were concatenated and fed into a fully-connected (FC) layer trained with a hinge loss. 
The performance was evaluated using the area under curve (AUC) metric. The multimodal architecture performed ca.\ 3\% better than the best unimodal network. Moreover, Gao et al.\ noted that the haptic classifier tends to have a high recall, predicting many adjectives for each class. In contrast, the visual classifier had higher precision. Finally, the multimodal classifier had higher precision and recall than the haptic classifier and higher recall than the visual classifier.

\cite{Tatiya2019Deep} implemented a \textit{post-mapping} fusion approach on the dataset by \cite{Sinapov2014Grounding} described in Section \ref{sec:datasets}. 
Tatiya and Sinapov applied a tensor-train gated recurrent unit (TT-GRU) for processing the visual information available in the dataset. Both the acoustic and haptic data in the dataset were processed using a CNN. For the acoustic data, the audio was preprocessed into two channels, the first consisting of the log-scaled Mel-spectrogram and the second of the spectrogram's derivative. 
The haptic data was downsampled from 500Hz to 50Hz to align with the video and acoustic data. 
The multimodal fusion network consisted of the concatenated output vectors of each unimodal network, a fusion layer, and an output layer. Each unimodal network was optimized to recognize the category of the objects. Thus, these networks can be used as stand-alone classifiers or integrated into a multimodal network. 
Overall, the results were comparable to the earlier work by \citet{Sinapov2014Grounding}. However, the baseline and the suggested approach have their strengths in different EPs data. Nevertheless, whether such complementary best performance can be attributed to the dataset or the architecture used is unclear.
\cite{Abderrahmane2018VisuoTactile} applied Zero-Shot Learning to an object classification task, in which a multimodal CNN trained on a set of objects was used to recognize novel objects that were never seen or touched before; relevant semantic attributes (e.g. round, soft, bumpy) were encoded from visuo-tactile data during training and then used to recognize the novel objects, with an accuracy of 72\%.
\cite{Taunyazov2020EventDriven} proposed a Visual-Tactile Spiking Neural Network (VT-SNN) that combines information coming from two event-driven sensors: a novel neuromorphic tactile sensor, NeuTouch, and a Prophesee event camera\footnote{https://www.prophesee.ai/}. The network was trained on two tasks: container classification and rotational slip detection. A comparative experimental analysis showed that the combination of vision and touch performed better than vision or touch alone.

\subsubsection{Transfer Learning}
\label{sec:transfer}
One of the challenges of transfer learning (co-learning) is that machine learning models are based on the assumption that both training and test data are drawn from the same distribution. 
However, such an assumption does not hold when transferring knowledge between different robots or sensor modalities. 
A possible solution is \textit{domain adaptation}, a.k.a.\ transfer learning, \cite[e.g.,][]{DaumeIII2006Domain, Wang2018Deep}. Here, training samples from a source dataset are adapted to fit a target distribution.

One example of \textit{domain adaptation} applied to multimodal object recognition was recently presented by \cite{Tatiya2020Framework}.
Tatiya et al.\ used a probabilistic variational auto-encoder network ($\beta$-VAE) to cope with missing or defective sensors or new behavioural modalities such as those related to a new exploration procedure. They also implemented a probabilistic variational encoder-decoder network ($\beta$-VED) to transfer knowledge from one or multiple robots to another. In both cases, the $\beta$-VAE and $\beta$-VED were implemented using multi-layer perceptrons, and object classification was performed using an SVM. 
For testing, the dataset of \cite{Sinapov2014Grounding} described in Section \ref{sec:datasets} was used. In particular, 15 of 20 object categories were randomly selected for training, and the five remaining were used to test transfer learning between sensory modalities or different behaviours.
Tatiya et al.\ report that such an approach based on $\beta$-VAE and $\beta$-VED can effectively transfer feature representations from one or more sensory modalities to another with a performance comparable to learning those representations from scratch.

\citet{Falco2019Transfer} presented a four-steps visual-to-tactile transfer architecture for object recognition.
Firstly, a visuo-tactile \textit{common representation} based on point clouds was preprocessed to obtain similarly sized representations. In particular, \textit{equalizing partiality} allowed to filter out the noise and reconstruct missing portions of the surface, and \textit{uniforming density} was used to downsample the point density while creating a more uniform point density. 

Secondly, despite preprocessing, the representation of both modalities is still imperfect. Thus, the redundancy of the information was increased to create a more robust \textit{feature set} which was later compressed using singular value decomposition (SVD). 

Thirdly, \textit{transfer learning} three methods based on dimensionality reduction were tested, namely, transfer component analysis (TCA), subspace alignment (SA), and geodesic flow kernel (GFK). TCA and SA learn feature representations that are invariant across domains. In contrast, GFK focuses on geometric and statistical changes from the source domain to the target domain. 

Finally, for \textit{object classification} k-NN and SVMs were compared. 
The architecture was tested with a dataset of 15 objects, including 40 visual and five tactile samples per object. The version using transfer learning based on GFK and an SVM achieved an accuracy of up to 94.7\%, comparable to classification results for unimodal object recognition in this dataset. Moreover, \citet{Falco2019Transfer} reported that the preprocessing step contributes about 13\% of the performance while the GFK transfer learning accounts for 20\% of the performance. 
The other transfer learning methods tested achieved a very low accuracy. A possible disadvantage of the proposed methods is the need for both the source data and (portion of) the target data.

\citet{Tatiya2020Haptic} proposed a framework for knowledge transfer using kernel manifold alignment (KEMA). 
Manifold alignment aligns datasets and projects them into a common latent space. The local geometry of each manifold is preserved while the correlations between manifolds are extracted. In KEMA, the common latent space was used for training instead of each robot's raw sensory data, allowing knowledge transfer.

Then two multi-class SVM models with the RBF kernel were trained, one dedicated to speeding up object recognition and the other to recognising novel objects.
For the first case of speeding up recognition, \citet{Tatiya2020Haptic} used two source robots with extensive experience of the objects and a novice robot with limited experience. The sensory experience of all three robots was used to build the latent space and train the model. The results showed a delicate balance between the amount of source data used and performance. However, when that balance was met, the target robot performed consistently better than a robot trained only using its own sensory data.

For the novel object recognition case, \citet{Tatiya2020Haptic} used two expert robots and a novice (target) robot having extensive experience with a few objects and no experience with other objects. The sensory data of all three robots were used to train the model. The results showed that KEMA could transfer existing knowledge to the target robot, accurately classifying all unseen objects. Different variations of the experiments showed that the target robot consistently achieved better than chance accuracy. 
Some of the limitations of this approach were the need to use the target robot's sensory data for training the model and the need for all robots to perform the same actions on the same objects. Another limitation was that all experiments were performed with simulated robots, and the only haptic difference was the objects' weight.

\cite{Luo2018ViTac} applied maximum covariance analysis (MCA) for crossmodal texture recognition. 
They introduced the ViTac dataset, consisting of 100 different cloth textures collected with an RGB camera and a GelSight sensor.
For MCA, both modalities were preprocessed independently. Then, these features were used to create a covariance matrix, and finally, singular value decomposition (SVD) was applied to reduce the dimensionality. 
MCA is typically used with handcrafted features to create the covariance matrix. However, Luo et al.\ used a pre-trained AlexNet, replaced the fully-connected layers, and called their method DMCA. 
Both visual and tactile data were presented durfing the learning phase. However, only one modality was used for testing. 
Luo et al.\ showed that the classification performance of DMCA improves as the output dimension increases, reaching a maximum performance at approximately 25 output dimensions. The classification performance for tactile data was ca.\ 90\%, while the classification performance for visual data was ca.\ 92.6\%. In both cases, these results were ca.\ 7\% better than the unimodal classification case in this data using a pre-trained AlexNet.

\cite{Lee2019Touching} presented conditional generative adversarial nets (cGANs) to generate visual data from tactile sensory input and vice-versa. They used the ViTac dataset of cloth textures, which consists of 100 different pieces of fabric. 
The dataset has RGB macro images of the fabrics and tactile readings from a GelSight sensor. The results showed that visual-to-tactile generation achieves a similarity of around 90\%. Whereas generation from tactile-to-visual achieved similarities ranging from 50\% to 90\%. Finally, the classification of both generated and original visual and tactile images achieved an accuracy of ca.\ 90\%. Data augmentation seemed to be a promising direction for some modalities, particularly from a higher dimensional modality like vision to a lower-dimensional one like tactile images.

\subsection{Multimodal Peripersonal Space Representation}
\label{sec:peripersonal}
The peripersonal space (space immediately surrounding the body) is crucial for effective interaction with the environment. Examples of work on this area are presented by \cite{Bhattacharjee2015Combining} in which an iterative algorithm is used to extrapolate haptic labels (force data) to regions of an RGB-D image with a similar colour and depth as those for which the haptic data was explicitly measured. The algorithm operates under the assumption that visible surfaces that look similar to one another are likely to have similar haptic properties. The algorithm can reach an average performance of 76.02\% employing 40 contact points in simulation. For haptic categorization, a Hidden Markov Model (HMM) based classification method was employed, which takes force data as input and outputs sparse haptic labels, each with a 2D colour image coordinate. Later, \cite{Shenoi2016CRF} used a dense Conditional Random Field (CRF) to produce a haptic map based on the HMM classification and a vision-based haptic label estimation using a CNN. This approach improved the average performance to 93\% for 40 contact points in the simulation. When tested on a foliage environment, the algorithm achieves 82.52\% performance after ten reaches.

A cognitive-inspired model for peripersonal space learning presented by \cite{Roncone2016Peripersonal} was implemented on the iCub robot. 
The model is used to learn approach/avoidance behaviour with the closest body part based on the distance and velocity of the stimuli. 
The model is fast to learn (a single interaction can already produce a functional representation which can be refined over time), capable of learning distributed representations incrementally, and stimuli agnostic. Thus, the algorithm can be used online and in real time without pretraining. 
The use of the distributed representation, although overall beneficial, imposes high computational and memory requirements. 
The current implementation assumes the robot's kinematics, and the different reference frames transformation is given. Other assumptions include the motor primitives used for learning (i.e., double-touch behaviour).
The model's implementation follows a developmental timeline. It is divided into three phases: starting with data collection through self-exploration or self-touch (motor-tactile stimulation), followed by data from external approaching objects considering time to contact (visuo-tactile stimulation). Finally, learning approach/avoidance behaviours irrespective of whether the perceived stimulus is of motor or visual origin.

Building upon \cite{Roncone2016Peripersonal}, \cite{Straka2017Learning} proposed a model using a Restricted Boltzmann Machine and a feedforward neural network. The stimulus's position and velocity are estimated visually and represented as a normal distribution to account for uncertainties. The resulting representation is then fed into a feedforward neural network that learns to predict a contact's location. The model was tested on a simulated 2D scenario and can expand the Peripersonal Space when confronted with fast stimuli. It can also confidently predict contact based only on visual estimations of position and velocity.

\subsection{Multimodal Object Perception for Manipulation}
\label{sec:manipulation}
Robotic manipulation has a huge impact in many industrial and service applications; visuo-tactile perception has been actively studied to improve the performance of robots, for instance, by allowing more secure object grasping and handling with a lower risk of damaging delicate objects.
In the multimodal setting, visual perception is predominantly used for planning reaching trajectories and identifying grasp type and orientation, while haptic perception is typically used for slippage prevention and compliant grasping. The classical way of tackling the problem of grasping has been with model-based, i.e., analytical approaches, and examples of such multimodal perception for grasping and manipulation in the literature are abundant. However, as seen in other fields, recently, there has been a tendency to move from model-based approaches to data-driven ones.
In this section, we outline the importance of using both the visual and haptic modality for grasping and manipulation tasks by presenting several recent approaches whose results show that multimodal variants are outperforming the uni-modal ones; see \cite{Bohg2014DataDriven} for an in-depth survey of older data-driven grasping approaches.

\subsubsection{Reaching}
\citet{Nguyen2019Reaching} proposed a visuo-proprioceptive-tactile integration model for a humanoid robot based on how infants learn to reach for an object. The authors used the iCub robot in simulation, with emulated tactile sensor regions distributed along the left arm and forearm representing the haptics modality, images from the two eye-cameras of the robot representing the visual modality and the configuration of the head, arm and torso joints representing proprioception. The proposed model uses the images from the eye-cameras and its head joints configuration as an input and predicts a list of the torso and arm joints configurations for reaching the object.
Convolutional feature extractors were used to extract feature descriptors from the visual input, after which the descriptors from both visual streams were concatenated with the head joints values. The concatenated descriptors were fed to a two-layer MLP, from which a third layer branched out to provide region-specific weights for mapping each of the 22 tactile regions to an input-specific arm-torso joint configuration. 
The trained model could successfully infer arm-torso configurations to perform region-specific reaching of the object with the arm or the forearm.

\subsubsection{Grasping}
Once an object is reached, the robot can grip the object and lift it. At this stage, it is crucial to find a good gripper configuration and to apply an adequate force such that the grasp is successful. \citet{Calandra2018More} presented a data-driven action-conditioned approach for predicting grasp success that can be used to determine the most promising grasping action based on raw visuo-tactile information.
Given an action consisting of 3D motion, in-plane rotation and change of force applied by the gripper, the proposed model uses a midst-mapping fusion strategy to combine the different modalities and predict the grasp outcome. First, the visual input from a Kinect v2 camera and the tactile input from two GelSight sensors attached to the fingers of a Weiss WSG-50 gripper are separately processed by CNNs, while an MLP processes the action channel. Then the latent features were concatenated and fed to an MLP that outputs the probability of successful grasp. The results show that the multimodal variant outperformed uni-modal or hard-coded baselines when grasping previously unseen objects. 
Furthermore, the qualitative analysis shows that the model learned meaningful grasping strategies for positioning the gripper and what amount of force to apply for successful grasping. 

In the same direction, \cite{Cui2020SelfAttention} suggested a visuo-tactile fusion learning method with a self-attention mechanism for determining the grasp outcome. Their model's architecture consists of three modules: a feature extraction module, a module incorporating visual-tactile fusion and self-attention, and a classification module predicting whether a grasp would be successful. The feature extraction modules for the vision and tactile channel are based on CNNs. The feature fusion module performs a slice-concatenation of the visual and tactile features of particular positions in the corresponding feature maps. Then the self-attention mechanism generates a weighted feature map that learns to determine the importance of different spatial locations. In this way, the overall architecture could learn some aspects of the cross-modal position-dependent features. Finally, the classification module, composed of two fully-connected layers, maps the extracted visuo-tactile features to either a successful or unsuccessful grasp. The authors ran experiments and ablation studies considering different model input variants and tactile signal types, reporting state-of-the-art results on two publicly available datasets.

\subsubsection{Maintaining Grasping}
Once the object is grasped and lifted, slip detection is essential for maintaining a successful grasp. For instance, the gripper force can be adjusted to prevent objects from dropping when a slip is detected. In this direction, \cite{Li2018Slip} proposed a data-driven visuo-tactile model for slip detection of grasped objects based on DNN architecture. Their model uses a sequence of eight consecutive GelSight and corresponding camera image pairs during a grasp-and-lift attempt. Each modality undergoes a separate feature extraction step through a pre-trained CNN, after which the latent features for both modalities are concatenated (midst-mapping) and passed through an additional FC layer. 
LSTM layers are used on top of the FC layer, and a final FC layer provides the probability that a slip occurred for the duration covered by the image sequence. During the experimental evaluation, several conditions were tested, like the type of image input (raw vs difference images), type of feature extractor (different off-the-shelf CNN models) or the type of information (visual, tactile or visuo-tactile). The best performing model used combined visuo-tactile information, significantly outperforming the unimodal approaches and achieving 88\% accuracy in detecting slips on a test dataset of unseen objects.

\subsubsection{Multi-stage Grasping Pipelines}
Unlike the previously mentioned end-to-end learning approaches, \cite{Ottenhaus2019VisuoHaptic} proposed a multi-stage pipeline to combine vision and haptic information for finding the most suitable grasp pose. Depth information of the object's front side and touch information from its backside were fused to construct a precise voxel representation of unknown objects. Next, planners proposed grasp hypotheses, for which a neural network provided scores to decide on the most suitable grasp. Finally, the \textit{approach} and \textit{grasp} actions to lift the object of interest were executed. 
While the authors used existing methods for different parts of the pipeline, their main contribution was the neural network that can propose grasp scores from the voxel representation of the object and the rotation matrix of a grasp pose candidate. 
The network architecture is an example of midst-mapping fusion, where the output of a CNN feature extractor for the voxel input and an MLP feature extractor for the pose input is concatenated and fed into a final MLP that predicts the probability of a successful grasp. The neural network was trained in simulation, but its performance was validated on a real ARMAR-6 humanoid robot, with a head-mounted Primesense RGB-D camera and a force-torque sensor in the wrist of the robot's arm used for haptics.

Another multi-stage pipeline was recently proposed by \cite{Siddiqui2021Grasp}. Firstly, RGB-D sensing from a Kinect V2 camera was used to identify an approximate object pose with a 3D bounding box; then, the motion of a UR5 robot arm was planned to bring a multi-fingered Allegro robot hand equipped with Optoforce fingertip force sensors near to the located object. Finally, a haptic exploration procedure was performed, in which the hand touched the object several times with different tentative grasps, without lifting it, while evaluating a force closure grasp metric at each attempt. The haptic exploration was realized with unscented Bayesian optimization to reduce the number of exploration steps \citep{Nogueira2016Unscented, Castanheira2018Finding}. Unscented Bayesian optimization outperformed both Bayesian optimization and random exploration, i.e., uniform grid search. Overall, this method permitted to find safe and robust grasps for unknown objects without needing any previous learning, but at the cost of requiring considerable time (i.e., in the order of minutes) to haptically explore the object before lifting it.

\subsubsection{Contact-rich manipulation}
While traditional robotic manipulation is all about avoiding physical contacts with the environment that surrounds the objects, human manipulation is to a large extent about exploiting those contacts, as noted by \cite{Deimel2016Exploitation}. Inspired by this observation, and by the presence of several applied example in industry, such as peg-in-hole insertion tasks \citep{Jiang2020StateoftheArt}, the robotics community is showing increased interest in the development of robotic solutions for contact-rich manipulation tasks, as summarised by \cite{Suomalainen2022Survey}. Clearly, visual perception is not enough for these tasks, and visuo-haptic integration becomes crucial.
As a most notable example, \cite{Lee2019Making} recently proposed a system in which a robotic manipulator learns by deep reinforcement learning a control policy that includes sensory feedback from visual (RGB camera), haptic (force/torque sensor) and proprioceptive (motor encoders) sensing. A shared and compact representation of the high-dimensional and heterogeneous multimodal data is learned with a neural network, which is trained to predict optical flow, presence of contact, and concurrency of visual and haptic data; the neural network is then used as sensory feedback to learn a control policy for a peg insertion task, directly on the real robot. The experiments compare four models: no sensory feedback, vision only, haptics only, vision and haptics. Interestingly, while the model with haptics only performs as bad as the one with no feedback, because the robot cannot even pick the peg in most trials, the model with vision only performs the insertion successfully only about 50\% of the times, while the model with both vision and haptics brings the success rate to about 75\%. 


\section{Discussion and Outlook}
\label{sec:outlook}
Visuo-haptic object perception is a vibrant and dynamic field whose development is crucial for new sensing technologies and applications such as robotic grasping, smart prostheses, and surgical robots. This article highlights many foci of ongoing research from the theoretically and biologically inspired approaches, passing via sensor technologies, data collection, and finally, data processing and applications. However, numerous crucial challenges need to be overcome. This section summarizes and discusses some of these challenges.

\subsection{Biologically-inspired Approaches}
Regarding biological inspiration, the question for robotics is which and in what proportion bio-inspired sensory and data processing principles can help us improve multimodal object recognition in its multiple application areas. 
Sensor technologies are largely bio-inspired, and there are efforts to incorporate other capabilities, such as measuring humidity, hardness, and viscosity, as well as mimicking other skin properties such as self-healing \citep{Oh2019Stretchable}.
On the contrary, perception models in artificial agents are still largely detached from their biological counterparts. While some biological principles have been explicitly studied, like \textit{integration strategies} \citep[e.g.,][]{Toprak2018Evaluating}, others like \textit{hierarchical processing} and \textit{input-driven self-organization} or processing of object properties rather than sensory modality are some of the promising directions that should be further explored.

\subsection{Sensor Technologies}
Tactile sensing technologies require advancements in several aspects before they can be deployed as easily as cameras. Advancements not limited to the following areas are needed: mechanical robustness, flexibility, compliance, a decrease in electrical connections, sensitivity and reliability of the measurements, the capability of detecting multiple contacts simultaneously, detectability of both normal and shear forces, affordability and ease of manufacturing, as well as ease of electromechanical integration and replacement.

\subsection{Data Collection and Datasets}
Collecting tactile data during grasping on a real robot or correctly simulating tactile sensors for synthetic data generation are resource-intensive tasks, which in turn is reflected in datasets' availability and size. While there are many large-scale vision-only datasets for grasping in real-world scenarios or simulation \cite[e.g.,][]{Jiang2011Efficient, Levine2018Learning, Depierre2018Jacquard}, only a few small-scale visuo-tactile datasets exist. Thus, large-scale multimodal datasets should be created, considering a variety of objects, grasping scenarios and different tactile sensor types. 
However, data acquisition from tactile sensors still \textbf{lacks a unified theoretical framework}. The challenges here stem from the fact that haptic perception is an intrinsically sequential process. Moreover, haptic perception is highly dependent on the robot's embodiment which makes the generalization to other robots or tasks difficult. In addition to a unified theoretical framework for data acquisition, solving other standing computational challenges such as \textit{representation learning}, \textit{mapping} and \textit{co-learning} seem to be key enabling technologies that could help cope with the resource-intensive nature of data acquisition.

Real-world tactile data collection will continue to be the most relevant, and it will also continue to be the most resource-intensive to obtain. In light of recent improvements in the simulation approaches \cite[]{Wang2022Tacto, Lin2022Tactile} that allow generating synthetic data from different tactile sensors or improve the sim2real transfer \cite[]{Josifovski2018Object, Jianu2022Reducing, Gao2022ObjectFolder, Josifovski2022Analysis} for visual, tactile or proprioceptive sensing, it is expected that synthetic data gains popularity. Although synthetic data might not be sufficient, it might be a valuable and effective way to move the field forward when combined with small-scale real-world datasets.

\subsection{Multimodal Signal Processing and Applications}
With regards to signal processing and applications, even though multimodal visuo-haptic approaches for grasping show better results and have the potential to handle use-cases where visual information alone is insufficient, vision-only grasping approaches \cite[e.g.,][]{Levine2018Learning, Mahler2017DexNet, Bousmalis2018Using, James2019SimtoReal} are still more popular. Some reasons for this popularity are that the availability, durability and understanding of vision sensors are better than tactile ones. Moreover, the simulation of vision sensors is easier and more realistic, and the collection, processing and interpretation of visual information are easier than tactile sensor readings. 
On the other side of the spectrum, there are also recent grasping approaches \cite[e.g.,][]{Murali2020Learning, Hogan2018tactile} that only use tactile information, but such approaches are usually only suitable for limited scenarios or parts of the grasping process. 

Thus, future efforts should be concentrated on multimodal approaches. However, as discussed by \citet{Xia2022Review}, the main challenge is ensuring safety during the physical contact between the object and the robot necessary for tactile sensing. To avoid the hardware dependencies and the safety risks, simulations are a promising alternative to real-world training and data collection for learning-based grasping approaches. However, due to the inaccurate nature of simulations, they cannot completely replace, but they can significantly reduce, the amount of real-world data needed. Finally, fine-tuning on the real system or sim2real techniques \cite[e.g.,][]{Ding2020SimtoReal, Narang2021SimtoReal} can help to bridge the simulation-to-reality gap. 

Another major problem of data-driven and end-to-end learning grasping approaches is that they require a vast amount of training data, in contrast to humans, who learn and generalize from very few examples. In this regard, future work should concentrate on improving the sample efficiency of the algorithms. One option is to include priors in the learning process, e.g., meaningful relations between tactile sensing regions can be incorporated into the model through graph-like structures, e.g., \cite{Garcia-Garcia2019TactileGCN}. 
Another option is combining model-based and model-free techniques for grasping or developing hierarchical and multi-stage approaches. An added benefit of such approaches is that they provide better control over the grasping process and increased interpretability of the model's behaviour, which is crucial for applications in industrial or collaborative environments alongside humans. Safety is of utmost importance in such environments, and integrating tactile sensors like robotic skin \citep{Pang2021CoboSkin} can help improve tasks like grasping, prevent injuries, and enable compliant robot control.

\section{Conclusion}
This article provides a holistic overview of the current state of visuo-haptic object perception for robotic applications. First, it covers the biological basis of multimodal object perception in humans. Second, it summarizes sensor technologies, data collection strategies, and datasets. Third, it introduces the main challenges of multimodal machine learning, focusing on visuo-haptics. Fourth, it presents an overview of different applications. Finally, it presents a detailed discussion of the above points and future research directions for each of them. 

Despite the substantial advancements in the understanding and development in all those areas, there are still many open challenges, from the role of biological inspiration in multimodal object perception, to material and mechatronic advances required for the development of better tactile sensing technologies, to the development of better multimodal signal processing methodologies.

Covering the entire field of visuo-haptics for both biological and artificial agents in a single article is difficult. Thus, despite not being exhaustive, the holistic approach to the field presented in this article should provide a unique perspective to the reader on the current state and most pressing challenges that need to be addressed to continue moving the field of visuo-haptic object perception in robotics and its different applications forward.

\backmatter

\bmhead{Supplementary information} Not applicable

\section*{Declarations}

\bmhead{Funding} Open Access funding provided by the Projekt DEAL (Open access agreement for Germany).
\bmhead{Conflict of interest/Competing interests} The authors declare that they have no conflict of interest.
\bmhead{Ethics approval} This article does not contain any studies with human participants or animals performed by any of the authors.

\bmhead{Open Access} This article is licensed under a Creative Commons Attribution 4.0 International License, which permits use, sharing, adaptation, distribution and reproduction in any medium or format, as long as you give appropriate credit to the original author(s) and the source, provide a link to the Creative Commons licence, and indicate if changes were made. The images or other third party material in this article are included in the article's Creative Commons licence, unless indicated otherwise in a credit line to the material. If material is not included in the article's Creative Commons licence and your intended use is not permitted by statutory regulation or exceeds the permitted use, you will need to obtain permission directly from the copyright holder. To view a copy of this licence, visit \url{http://creativecommons.org/licenses/by/4.0/}.

\bmhead{Consent to participate} Not applicable
\bmhead{Consent for publication/Informed consent} Not applicable
\bmhead{Availability of data and materials} Not applicable
\bmhead{Code availability} Not applicable

\bibliography{references.bib}

\bmhead{Publisher's Note} Springer Nature remains neutral with regard to jurisdictional claims in published maps and institutional affiliations.

\end{document}